\PassOptionsToPackage{prologue,dvipsnames}{xcolor}
\documentclass[sigconf]{acmart}

\AtBeginDocument{%
  }

\copyrightyear{2024}
\acmYear{2024}
\setcopyright{acmlicensed}\acmConference[MM '24]{Proceedings of the 32nd ACM International Conference on Multimedia}{October 28-November 1, 2024}{Melbourne, VIC, Australia}
\acmBooktitle{Proceedings of the 32nd ACM International Conference on Multimedia (MM '24), October 28-November 1, 2024, Melbourne, VIC, Australia}
\acmDOI{10.1145/3664647.3680838}
\acmISBN{979-8-4007-0686-8/24/10}

\usepackage{caption}

\usepackage{colortbl}
\usepackage[ruled,linesnumbered]{algorithm2e}

\usepackage{amsmath}
\usepackage{graphicx}
\usepackage{multirow}
\usepackage{listings}
\usepackage{comment}
\usepackage[utf8]{inputenc}
\usepackage[T1]{fontenc}

\DeclareUnicodeCharacter{FF0C}{,}
\begin{document}

\title[Advancing Multimodal Large Language Models with Quantization-Aware Scale Learning \\ for Efficient Adaptation]{Advancing Multimodal Large Language Models with Quantization-Aware Scale Learning for Efficient Adaptation}

\author{Jingjing Xie}
\affiliation{%
  \institution{Key Laboratory of Multimedia Trusted Perception and Efficient Computing, Ministry of Education of China, Xiamen University}
   \city{Xiamen}
   \country{China}
 }

\author{Yuxin Zhang}
\affiliation{%
  \institution{Key Laboratory of Multimedia Trusted Perception and Efficient Computing, Ministry of Education of China, Xiamen University}
   \city{Xiamen}
   \country{China}
 }

\author{Mingbao Lin}
\affiliation{%
  \institution{SkyWork AI}
   \country{Singapore}
 }

\author{Liujuan Cao}
\authornote{Corresponding author (caoliujuan@xmu.edu.cn)}
\affiliation{%
  \institution{Key Laboratory of Multimedia Trusted Perception and Efficient Computing, Ministry of Education of China, Xiamen University}
   \city{Xiamen}
   \country{China}
 }

\author{Rongrong Ji}

\affiliation{%
  \institution{Key Laboratory of Multimedia Trusted Perception and Efficient Computing, Ministry of Education of China, Xiamen University}
   \city{Xiamen}
   \country{China}
 }



\begin{abstract}
This paper presents the first study to explore the potential of parameter quantization for multimodal large language models to alleviate the significant resource constraint encountered during vision-language instruction tuning.
We introduce a Quantization-aware Scale LeArning method based on multimodal Warmup, termed QSLAW.
This method is grounded in two key innovations: (1) The learning of group-wise scale factors for quantized LLM weights to mitigate the quantization error arising from activation outliers and achieve more effective vision-language instruction tuning; (2) The implementation of a multimodal warmup that progressively integrates linguistic and multimodal training samples, thereby preventing overfitting of the quantized model to multimodal data while ensuring stable adaptation of multimodal large language models to downstream vision-language tasks.
Extensive experiments demonstrate that models quantized by QSLAW perform on par with, or even surpass, their full-precision counterparts, while facilitating up to 1.4 times reduction in VL tuning time and GPU consumption.
Our code is released at \url{https://github.com/xjjxmu/QSLAW}
\end{abstract}



\maketitle

\section{Introduction}

The remarkable performance of large language models (LLMs) has been well-established in recent literature~\cite{chen2020big,chowdhery2023palm,JMLR:v21:20-074,radford2019language,touvron2023llama}, sparking a growing interest in the development of multimodal large language models (MLLMs)~\cite{alayrac2022flamingo,liu2024visual,li2022blip,wu2023visual,team2023gemini,chen2023pali,luo2023cheap}. This burgeoning field has led to substantial progress in a wide array of vision-language (VL) tasks. To accomplish this, contemporary MLLMs primarily utilize multimodal instruction following examples for VL instruction tuning and adopt modular architectures~\cite{alayrac2022flamingo,liu2024visual,li2022blip,jian2024bootstrapping} to transform visual features into the word embedding space of the LLM. This innovative approach enables LLMs to execute multimodal tasks in an autoregressive fashion.
One notable example of this technique is LLaVA~\cite{li2022blip}, which employs a linear projection layer to bridge the gap between the visual encoder and the LLM. By doing so, LLaVA fully harnesses the power of pre-trained LLMs, thereby significantly enhancing its visual comprehension capabilities. 

Despite the advancements, the current VL instruction tuning for MLLMs exhibits considerable redundancy in terms of computation and memory burden. This limitation primarily stems from the inherently large size of LLMs compared to other components within MLLM architectures. For instance, LLaVA-13B fully fine-tunes the entire LLM during VL instruction tuning, often requiring hundreds of GPU hours~\cite{liu2024visual}. Although recent efforts have introduced more efficient adapters and the freezing of LLMs to reduce training overheads~\cite{luo2023cheap, Fang_2023_ICCV}, VL tuning within current MLLM frameworks still demands substantial memory usage and computational resources, necessitating at least 8 NVIDIA Tesla A100 GPUs~\cite{luo2023cheap}. This poses great challenges to the rapid adaptation of LLMs for cross-modal tasks, particularly in situations characterized by limited training resources and needs for on-the-fly, task-specific tuning.

To address this constraint, this paper explores the potential of parameter quantization for MLLMs, aiming to alleviate the extensive training demands encountered during VL instruction tuning while preserving the original performance. Quantization, a network compression technique, transforms the full-precision weights into low-bit representations, consequently reducing both computational load and storage requirement. 
It has been adopted for parameter-efficient fine-tuning (PEFT) of LLMs~\cite{guo2024lqlora,xu2024qalora,dettmers2024qlora}, notably in QLoRA~\cite{dettmers2024qlora}, which quantizes each linear layer's weights into a 4-bit NormalFormat (NF) datatype and uses the low-rank adapter (LoRA)~\cite{hu2022lora} for fine-tuning. Owing to the lightweight quantized LLM and a minimal set of trainable parameters within the LoRA module, QLoRA can facilitate LLaMA-65B fine-tuning on a single 48GB GPU without sacrificing chat performance~\cite{dettmers2024qlora}.
\begin{table}[!t]
    \centering
    \setlength{\abovecaptionskip}{0.3cm}
    \setlength{\belowcaptionskip}{-0.4cm}
    \caption{Cost and accuracy over various VL instruction tuning paradigms on ScienceQA. The symbol ``\dag'' denotes advanced memory-saving strategies, while ``OOM'' indicates GPU memory exhaustion. Results are evaluated using 4 A800 GPUs.}
    \resizebox{\linewidth}{!}{
    \begin{tabular}{lcccc}
    \toprule
    \text { Methods } & \text { \#T-Params } & \text { Memory (GB) } & \text { Time (hours) } & \text { Average (\%) } \\
    \midrule
\text{ LLaVA~\cite{liu2024visual}} & \text{13}B &  OOM &  \text { N/A } & N/A  \\
\text{ LLaVA$\dag$~\cite{liu2024visual}} & \text{13}B &  71.54 &  \text {8.75} & 90.92  \\
    \text { QLoRA } &  \text{500.70}M &  66.92 &  \text { 6.12 } & 86.96\\
    \text { QSLAW } &   \text{84.25}M & 66.52  &  \text { 5.76 } & \textbf{91.04} \\
    \bottomrule
    \end{tabular}
}
    \label{tab:cost-acc}
\vspace{-0.5cm}
\end{table}

\begin{figure}[!ht]
    \centering
    \setlength{\abovecaptionskip}{0.3cm}
    \setlength{\belowcaptionskip}{-0.cm}
    \includegraphics[width=0.45\textwidth,height=0.3\textheight, keepaspectratio]{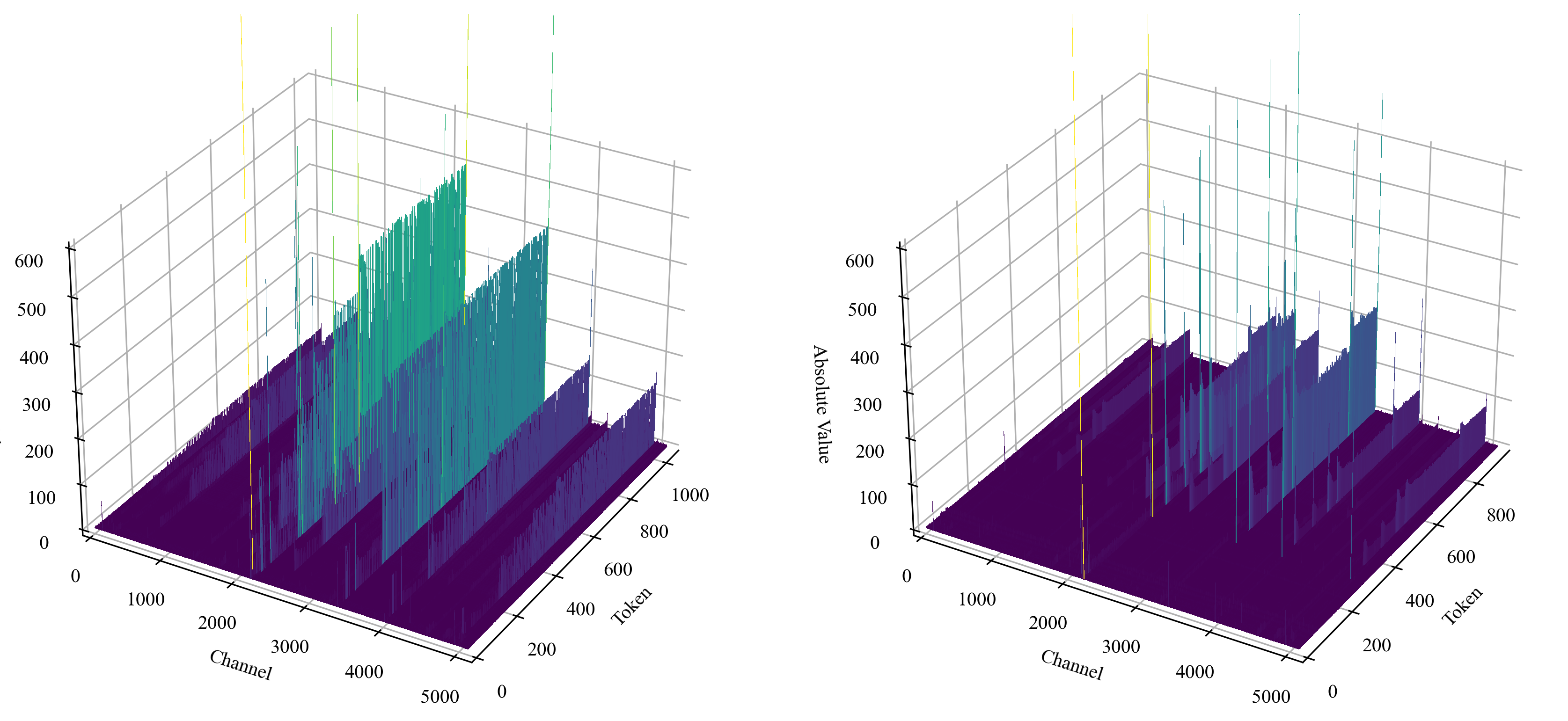}
 \caption{Absolute magnitude of the input activation in one LLaVA-13B block. Left (image and text tokens) exhibits a larger scale in activation compared to the right (only text tokens).}
 \Description{activation outlier of language-only and multimodal data.}
    \label{fig:outlier}
    \vspace{-2em}
\end{figure}

A potential strategy to consider is implementing the previously discussed PEFT method to facilitate VL instruction tuning for MLLMs. We conduct an experiment in Table\,\ref{tab:cost-acc} to analyze its efficacy.
As can be seen, utilizing QLoRA to quantize LLM weights to 4-bit can significantly reduce both GPU memory consumption and time overhead. Regrettably, QLoRA inflicts a considerable performance impairment on multimodal tasks, with almost a 4\% accuracy decrease on ScienceQA~\cite{lu2022learn}, despite its capacity to attain parity with full-precision performance in language tasks~ \cite{hendrycks2020measuring,wang2018glue}.
This incongruity prompts us to investigate the effects of quantization on MLLMs during VL instruction tuning in greater depth. Consequently, we examine the activation distribution within intermediate layers of LLMs, focusing on language and multimodal image data, as depicted in Figure\,\ref{fig:outlier}.
A noticeable percentage of activations emerge as outliers, displaying significant deviations in magnitude, which poses a substantial challenge for MLLMs quantization. Minor quantization errors may accumulate and interact with these activation outliers, ultimately resulting in irreversibly distorted outputs~\cite{dettmers2022gpt3,lin2023awq}. 
Furthermore, the density and frequency of activation outlier markedly increase in multimodal inputs compared to unimodal language inputs. This observation elucidates the performance deterioration of QLoRA in VL instruction tuning for MLLMs, as its adopted NF4 datatype only pursues equating the quantity of values across all quantization bins from the weight tensor and causes severe information loss, 
making it hard for LoRA to accommodate activation outliers.

To address this limitation, we employ quantization-aware scale learning instead of using LoRA to fine-tune the quantized MLLLM.
Specifically, we divide the weights into multiple quantization groups, each assigned a learnable scale factor.
This scale learning approach effectively reduces quantization errors within each group, particularly in cases where activations exhibit outlier characteristics at certain positions.
Furthermore, we adopt uniform quantization instead of forcing an equal number of weights in each quantization bin like NF4 and initialize the quantized weights with OmniQuant~\cite{shao2024omniquant}, an LLM uniform quantization method that employs weight clipping to mitigate the quantization difficulty occurring in language tasks, rather than relying solely on the probability density of the weights as in NF4.
Compared to LoRA, our proposed scale learning method offers several distinct advantages.
First and foremost, LoRA fine-tuning targets the output of the entire layer, which fails to adaptively minimize the quantization errors at outlier positions.
Additionally, scale learning exhibits significantly higher efficiency than LoRA.
For example, with a group size of 128, the introduced parameters for scale learning amount to only 16.83\% of those required for fine-tuning with LoRA, thereby enabling more efficient VL instruction tuning for MLLMs.

Next, we explore the data construction for scale learning. 
Relying solely on multimodal data for training scaling tends to cause the LLM to overfit to downstream tasks, subsequently diminishing its inherent linguistic proficiency. 
On the other hand, using a heterogeneous mix of language and multimodal data compromises the efficiency of VL tuning, as MLLMs fail to receive adequate multimodal instructional guidance during the initial stages of training. 
To address this issue, we introduce a novel modality-aware warmup method, which utilizes only multimodal data during the early phase of VL tuning and subsequently incorporates language data for scale learning. 
This ensures that MLLMs receive precise multimodal instructional supervision and avoids potential overfitting of the LLM backbone on multimodal data in the later stages of training, thereby preserving its original linguistic knowledge.

Our Quantization-aware Scale LeArning based on multimodal Warmup, termed QSLAW, is demonstrated to be effective for efficient MLLM instruction tuning across various VL tasks. For instance, QSLAW achieves 91.04\% accuracy on ScienceQA with LLaVA-13B, representing a 4.08\% gain compared to the 86.96\% achieved by QLoRA, which even outperforms the full-precision fine-tuned LLaVA-13B (91.04 for QSLAW \emph{v.s.} 90.92 for full-precision). Our contributions include:
\begin{itemize}
    \item We undertake the pioneering exploration of MLLMs quantization and utilize scale learning to alleviate the quantization challenges arising from the frequent occurrence of activation outliers inherent to MLLM quantization.
    \item We introduce a modality-aware warmup called multimodal warmup to prevent the quantized model from overfitting to multimodal data while ensuring stable adaptation of MLLMs to downstreams.
    \item Extensive experiments validate that QSLAW significantly reduces training time and memory footprint for VL instruction tuning while maintaining state-of-the-art performance.
\end{itemize}

\section{Related Work}
\subsection{Model Quantization}

Quantization methods can be broadly classified into quantization-aware training (QAT) and post-training quantization (PTQ). QAT relies on the complete training data to fine-tune the quantized model in line with the pre-training phases~\cite{esser2019learned,liu2023llm,choi2018pact,shin2023nipq,liu2022nonuniform}.
Although QAT exhibits promising performance retention, it
%
suffers the need for training weights and quantization parameters on the full dataset. 
In contrast, PTQ~\cite{nagel2020u,li2021brecq,wei2022qdrop} can efficiently perform quantization with significantly less data and resources.

Although these methods are highly efficient for CNNs, they cannot be extended to LLMs due to the difficulty of optimizing the vast parameter space with limited samples. GPTQ~\cite{frantar2023optq} is the first attempt to implement PTQ on models with billions of parameters, utilizing second-order information to compensate for quantization error. LLM.int8()~\cite{dettmers2022gpt3} highlights significant errors in quantization caused by activation outliers within LLMs, leading to the proposal of a mixed-precision quantization approach. AWQ~\cite{lin2023awq} proposes input channel scaling based on activation to protect essential weights. OmniQuant~\cite{shao2024omniquant} designs learnable weight clipping and learnable equivalent transformation, making them differentiable to quantize LLMs with gradient optimization. AffineQuant~\cite{ma2024affinequant} further proposes directly optimizing equivalent affine transformations. These methods aim to reduce the memory footprint of LLMs during inference. QLoRA~\cite{guo2024lqlora} is the first to explore reducing memory footprint during training through quantization and LoRA~\cite{hu2022lora}. LoftQ~\cite{li2024loftq} and LQ-LoRA~\cite{guo2024lqlora} alternate between quantization and singular value decomposition to find a suitable initialization for LoRA. QA-LoRA~\cite{xu2024qalora} proposes balancing degrees of freedom between quantization and adaptation with group-wise down projection of LoRA. These methods aim to enhance LoRA based on the NF data type, contributing orthogonally to our approach. Our method falls within the realm of quantization for LLMs. Unlike the aforementioned methods, we specifically address the quantization challenges in MLLMs, where additional visual inputs can influence the activation distribution and pose a new challenge. Also, we are the first to explore quantization challenges for MLLMs.

\subsection{Multimodal Large Language Models}

Traditional VL instruction tuning commonly employs various task-related losses, including image-text contrastive loss, image-text matching loss, and language modeling loss, to supervise the training of both visual and language branches. To compute these losses, it is typically necessary to perform multiple forward passes on the image-text pairs, consuming thousands of GPU hours. However, with the emergence of large language models (LLMs), the paradigm of VL tuning has shifted towards treating LLMs as a universal interface and adopting a modular structure to align representations from vision with LLMs. In these approaches, LLMs and the modular structure are trained on multimodal examples using a simple cross-entropy loss.
Recent advances in this area include Flamingo~\cite{alayrac2022flamingo}, which introduces the Perceiver Resampler as the modular structure, and BLIP2~\cite{li2022blip}, which proposes a lightweight Q-Former to align different modalities. LLaVA~\cite{liu2024visual} employs a simple MLP as a modular structure and introduces VL instruction tuning, enabling LLMs to execute multimodal tasks in an autoregressive manner.
Despite these advancements, the current VL instruction tuning for massive MLLMs remains expensive. For example, LLaVA-13B fully fine-tunes the entire LLM during VL instruction tuning, often requiring hundreds of GPU hours. LaVIN~\cite{luo2023cheap}, which utilizes an adapter to achieve parameter-efficient VL instruction tuning, still necessitates at least eight NVIDIA Tesla A100 GPUs.
Our method is designed to alleviate the extensive training demands in VL instruction tuning while preserving the original performance. And we are the first to achieve this for VL instruction tuning by employing quantization with a minimal set of trainable parameters.

\section{Methodology}
\subsection{Preliminary}\label{sec:preliminary}

The objective of vision-language (VL) instruction tuning for MLLMs is to adapt an LLM backbone from processing unimodal text data to encompassing multimodal data. Specifically, given a multimodal instruction following example that consists of an image $\mathbf{I} \in \mathbb{R}^{h \times w \times 3}$ and a text sequence $\mathbf{T} \in \mathbb{R}^{l}$, the image $\mathbf{I}$ is initially fed into an image encoder, typically a pre-trained vision transformer~\cite{dosovitskiy2021an}, to extract the informative visual representation as:
\begin{equation}\label{eq:imgenc}
    {\textbf{F}_\textbf{I}} = {f_{\theta_{I}}(\textbf{I})},
\end{equation}
where $\theta_I$ represents the encoder's parameters.
Then, the visual representation is projected to the word embedding space of LLMs through a modular structure parameterized by $\theta_a$:
\begin{equation}\label{eq:align}
    {\textbf{F}_\textbf{I}}' = {f_{\theta_{a}}(\textbf{F}_\textbf{I})}.
\end{equation}

Subsequently, the LLM with pre-trained weights $\mathbf{W}$ receives the embedded image feature and the text sequence $\textbf{T}$ to generate a probability distribution $\textbf{P} \in \mathbb{R}^{L\times N}$ for each word in the target response $\textbf{R}\in\mathbb{R}^L$:
\begin{equation}\label{eq:llm}
    {\textbf{P}} = {g_\mathbf{W}}{({\textbf{F}_\textbf{I}}', \textbf{E}_\textbf{T})},
\end{equation}
where $\textbf{E}_\textbf{T}=f_{\theta_{T}}(\textbf{T})$ is the word embedding of the input text sequence and $N$ denotes the vocabulary size of the pretrained LLM.

Finally, the modular structure and LLM are jointly fine-tuned by minimizing the cross-entropy loss, which can be formulated as:
\begin{equation}\label{loss}
    \mathcal{L} = -\sum_{i=1}^{L}{\log{\textbf{P}_{i,j}}},
\end{equation}
where $j$ represents the position of $\textbf{R}_i$ in the vocabulary.

Albeit the efficacy, it requires considerable computational resources and memory usage, mainly streaming from the significantly large parameters in LLMs. Although recent advancements~\cite{luo2023cheap,dettmers2024qlora,zhang2023llama,lester2021power} have shown the potential of freezing the LLM backbone to eliminate partial backward costs of the LLM backbone, existing VL tuning frameworks still require a minimum of 8 NVIDIA Tesla A100 GPUs. This necessity poses significant challenges to the efficient adaptation of LLMs to cross-modal tasks, particularly in situations characterized by limited training resources and the need for on-the-fly, task-specific fine-tuning.

\subsection{The Potential of LLMs Quantization}

Quantizing the parameters of an LLM backbone into lower-bit representations offers a promising solution to the above problem and has shown remarkable efficacy in traditional unimodal PEFT scenarios for LLMs~\cite{dettmers2024qlora,guo2024lqlora,li2024loftq,kim2024memory}. Therefore, we initiate an exploratory investigation into the potential of quantization for MLLMs, starting with a trial using the QLoRA~\cite{dettmers2024qlora} in PEFT contexts.

Specifically, QLoRA compresses the normalized weight $\hat{\mathbf{W}}$ into quantized weight $\overline{\textbf{W}}$ with 4-bit NormalFloat (NF) format $q_i$ as:

\begin{equation}\label{eq:quant}
   \overline{\mathbf{W}} = {\mathop{\arg\min}\limits_{q_i}}{|\hat{\mathbf{W}}-q_i|},
\end{equation}
where $\hat{\mathbf{W}} = \frac{\mathbf{W}}
{max(abs(\mathbf{W}))}$ and $q_i$ in 4-bit NF format is:
\begin{equation}\label{eq:nf}
   q_i ={\frac{1}{2}}(\,Q(\frac{i}{17})+Q(\frac{i+1}{17})\,),
\end{equation}
where $Q(\cdot)$ is the quantile function of the standard Gaussian distribution $\mathcal{N}(0,1)$. After the NF4 initialization, the quantized LLM is frozen, and the low-rank adapter (LoRA)~\cite{hu2022lora} is then utilized for transfer learning on downstream tasks. It has been widely demonstrated in the literature~\cite{zadouri2024pushing,chen2024longlora,hayou2024lora+,zhang2022adaptive} that QLoRA can maintain the performance of LLMs on downstream language tasks with 4-bit quantization, significantly reducing the training burden and memory costs. Regrettably, when we employ QLoRA to enhance the efficiency of MLLM training, a nearly 4\% decrease in accuracy is observed on ScienceQA, as illustrated in Table\,\ref{tab:cost-acc}. Upon analysis, the NF format described in Eq.\,(\ref{eq:nf}) only ensures an equal quantity of weight values across each quantization bin but overlooks activation outliers, which are common in LLMs. Furthermore, our observation in Figure\,\ref{fig:outlier} indicates that the density and frequency of activation outliers significantly increase with multimodal inputs compared to unimodal language inputs. Consequently, minor quantization errors may accumulate and interact with these activation outliers, ultimately leading to irreversible output distortion. In summary, although quantization holds considerable potential for alleviating the massive burden of MLLM tuning, the need to address the dense activation outlier phenomenon in multimodal scenarios remains pressing.

\begin{table*}
    \centering
        \setlength{\abovecaptionskip}{0.3cm}
\setlength{\belowcaptionskip}{-0.cm}
    \caption{Quantitative accuracy on ScienceQA test dataset. Question classes: NAT = natural science, SOC = social science, LAN = language science, TXT = text context, IMG = image context, NO = no context, G1-6 = grades 1-6, G7-12 = grades 7-12. The symbol ``\dag'' denotes a larger rank used for LoRA and the best results in each class are \underline{underlined}.}
    \label{tab:sqa}
\begin{tabular}{c|c|c|c|c|c|c|c|c|c}
\toprule
\multirow{2}{*}{ Method } & \multicolumn{3}{c|}{ Subject } & \multicolumn{3}{c|}{ Context Modality } & \multicolumn{2}{c|}{ Grade } & \multirow{2}{*}{ Average } \\
& NAT & SOC & LAN & TXT & IMG & NO & G1-6 & G7-12 & \\
 \midrule 
 \multicolumn{10}{l}{ \textit{\textbf{Zero-shot \& few-shot} representative methods with performance reported in the literature }} \\
 Human~\cite{lu2022learn} & 90.23 & 84.97 & 87.48 & 89.60 & 87.50 & 88.10 & 91.59 & 82.42 & 88.40 \\
 GPT-3.5~\cite{lu2022learn} & 74.64 & 69.74 & 76.00 & 74.44 & 67.28 & 77.42 & 76.80 & 68.89 & 73.97 \\
 GPT-3.5 w/ CoT~\cite{zhang2023llama} & 75.44 & 70.87 & 78.09 & 74.68 & 67.43 & 79.93 & 78.23 & 69.68 & 75.17 \\
\midrule 
\multicolumn{10}{l}{ \textit{\textbf{Two-stage} representative methods with performance reported in the literature} } \\
 LLaVA-13B ~\cite{liu2024visual} & 90.36 & 95.95 & \underline{88.00} & 89.49 & 88.00 & \underline{90.66} & 90.93 & \underline{90.90} & 90.92 \\
 LLaVA-13B-QLoRA & 74.20 & 79.19 & 69.55 & 74.10 & 70.40 & 72.47 & 75.51 & 71.39 & 74.04 \\
 LLaVA-13B-QLoRA\dag & 85.48 & 93.59 & 84.64 & 84.56 & 83.94 & 86.90 & 87.96 & 85.17 & 86.96 \\
 LLaVA-13B-QSLAW (Ours) & 83.26 & 91.79 & 80.00 & 82.80 & 81.30 & 83.07 & 86.01 & 80.95 & \textbf{84.20} \\
 LLaVA-13B-QSLAW\dag (Ours) & \underline{91.30} & \underline{96.06} & 86.45 & \underline{90.22} & \underline{89.34} & 89.34 & \underline{91.63} & 89.98 & \textbf{91.04} \\
\midrule
\multicolumn{10}{l}{ \textit{\textbf{One-stage} representative methods with performance reported in the literature}} \\
 LLaMA-Adapter~\cite{zhang2023llama} & 84.37 & 88.30 & 84.36 & 83.72 & 80.32 & 86.90 & 85.83 & 84.05 & 85.19 \\
 LaVIN-7B~\cite{luo2023cheap} & 89.25 & \underline{94.94} & 85.24 & 88.51 & 87.46 & 88.08 & 90.16 & \underline{88.07} & 89.41 \\
 LaVIN-7B-QLoRA & 87.61 & 94.04 & 85.18 & 86.51 & 85.62 & 88.43 & 89.02 & 83.08 & 88.27 \\
 LaVIN-7B-QSLAW (Ours) & \underline{90.23} & 93.59 & \underline{85.82} & \underline{89.54} & \underline{87.75} & \underline{88.71} & \underline{90.75} & \underline{88.07} & \textbf{89.79} \\
\bottomrule
\end{tabular}
\end{table*}

\subsection{Quantization-Aware Scale Learning}

We formally present our Quantizration-aware Scale LeArning based on multimodal Warmup (QSLAW) method, specifically designed for efficient Visual Language (VL) instruction tuning in MLLMs.
QSLAW addresses the challenges associated with MLLMs quantization from two aspects: (1) it uniquely learns scale factors for different weight groups, reducing the quantization error resulting from activation outliers and demonstrates to be more effective for VL instruction tuning on quantized LLMs; (2) it employs a modality-aware warmup strategy called multimodal warmup, which blends linguistic and multimodal training samples, thus preventing the quantized model from overfitting to multimodal data while ensuring a stable adaptation to the target VL tasks.

\textbf{Quantization-Aware Scale Learning.} 
During the VL instruction tuning process, we assign learnable group-wise scale factors $\textbf{s}$ to the LLM weights $\textbf{W}$ as follows:

\setlength{\abovedisplayskip}{3pt} 
\setlength{\belowdisplayskip}{2pt}
\begin{equation}
\label{eq:scale_factor}
    \hat{\mathbf{W}} = \frac{\mathbf{W}}{{\textbf{s}}}
\end{equation}

And then uniform quantization is utilized to convert $\hat{\mathbf{W}}$ into pseudo-quantized weight $\tilde{\textbf{W}}$:
\begin{equation}\label{eq:uniform}
    \tilde{\textbf{W}}=\textbf{\(\Delta \)}\times(\text{clamp}({\lfloor{\frac{\hat{\mathbf{W}}}{\textbf{\(\Delta \)}}}\rceil}+zp, 0, 2^k-1))-zp,
\end{equation}
where $\lfloor{\cdot}\rceil$ denotes the round-to-nearest integer operation and $k$ represents the quantization bit. \(\Delta \) and $zp$ are the quantization step-size and zero-point, respectively.

In this approach, $\textbf{W}$ is divided into multiple groups, with each group of weights being scaled by a single factor.
By learning the scale factor under the guidance of Eq.\,(\ref{loss}), the quantization error within each weight group can be effectively minimized towards downstream tasks, particularly for groups containing activations outlier.
To clarify, an appropriate scaling factor allows for the rescaling of weights into a quantile range that reduces the output perturbation when interacting with activation outlier exhibiting significant deviation magnitudes and is more suitable for quantized LLMs to transfer into VL tasks with VL instruction tuning.
In contrast, LoRA is unable to effectively mitigate such quantization errors caused by activation outliers, as it conducts fine-tuning in a coarse-grained, global manner.
Importantly, as demonstrated in Table\,\ref{tab:cost-acc}, the parameter count of the scale learning is substantially lower than that of the LoRA module, making our scale learning suitable for an efficient VL instruction tuning.

\begin{figure}[!t]
    \centering
    \setlength{\abovecaptionskip}{0.2cm}
    \setlength{\belowcaptionskip}{0.2cm}
    \includegraphics[width=\linewidth]{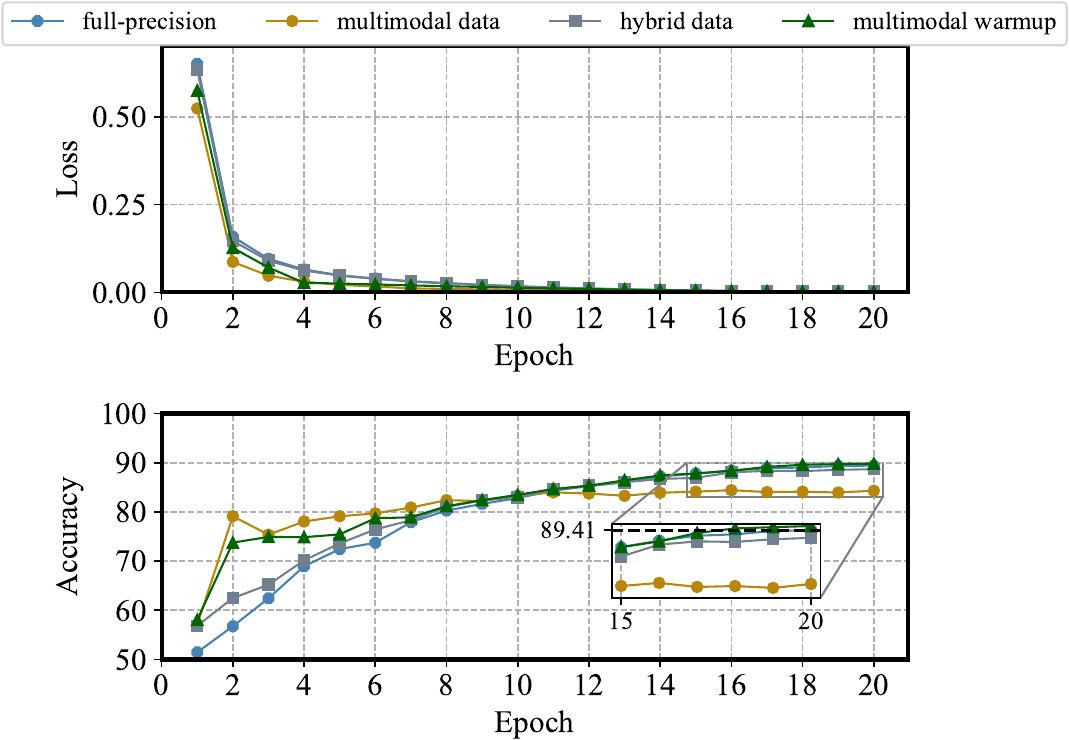}

    \caption{Loss and accuracy curves of training scaling with different strategies on ScienceQA. Solely utilizing multimodal data for training scaling tends to lead the LLM overfitted to downstream tasks. This is evidenced by a rapid decrease in loss but the accuracy remain mediocre.}
    \Description{Loss and Accuracy.}
    \label{fig:overfit}   
    \vspace{-0.7cm}
\end{figure}

\textbf{Modality-aware Warmup.} As discussed in Sec.\,\ref{sec:preliminary}, MLLMs fine-tune their parameters using instruction samples that encompass both images and textual content. Unfortunately, we find that training the scale factor for quantization using the same dataset can result in an overfitting issue for the LLM backbone. As illustrated in Figure\,\ref{fig:overfit}, scale learning on purely multimodal data leads to a rapid decrease in loss while the final model accuracy, paradoxically, fails to outperform the full-precision counterpart. This overfitting phenomenon is understandable, given that the LLM's pre-training was solely based on linguistic data. Consequently, conducting quantization-aware scale training exclusively on multimodal data can impair the inherent linguistic capabilities of the LLM, which are of paramount importance to serve as a language backbone for multimodal adaptation.

An intuitive solution to this overfitting problem involves the integration of linguistic data to jointly guide the scale learning process. Consequently, we supplement the existing multimodal data with the WikiText dataset~\cite{merity2016pointer}, thereby creating a hybrid dataset specifically designed for scale learning. Figure\,\ref{fig:overfit} illustrates the trajectories of loss and accuracy. While effective mitigation of overfitting is observed, the adaptive performance of the multimodal approach still falls short when compared to its full-precision counterpar. We attribute this outcome to an early-stage underfitting of the LLMs with respect to the multimodal data. Specifically, the MLLMs parameters, such as the scale factor and modular structure, are randomly initialized at the start of training, and the interference of linguistic supervision at this stage hinders the model's ability to fit the multimodal data, resulting in suboptimal performance.

To address the complex interplay between overfitting and underfitting, we equip QSLAW with a multimodal warmup data sampling strategy. More specifically, during the initial $\eta$ iterations of VL instruction fine-tuning, we exclusively utilize multimodal data pairs for scale learning. Subsequently, we incorporate linguistic text sequences extracted from the WikiText dataset to facilitate hybrid-data training. This warmup approach ensures accurate multimodal instructional supervision for the MLLMs during the initial training iterations, while simultaneously circumventing potential overfitting of the LLM backbone on multimodal data to preserve its inherent linguistic capabilities. Consequently, as demonstrated in Figure\,\ref{fig:overfit}, the proposed multimodal warmup method effectively rivals the accuracy of quantization-aware scale learning compared with its full-precision counterpart.

In addition to the two components mentioned above, our proposed QSLAW method initializes \(\Delta \) and $zp$ in Eq.\,(\ref{eq:quant}) using OmniQuant~\cite{shao2024omniquant}, a post-training quantization method for LLMs that employs weight clipping to mitigate the quantization challenge in language tasks, rather than solely relying on the probability density of the weights as in NF4. And it is worth mentioning that our method is not constrained by the quantization method or initialization and can achieve consistent performance improvements compared with LoRA as discussed in Sec.\,\ref{sec:init}.
%
%
Moreover, it is of crucial importance to realize that QSLAW is orthogonal to most of off-the-shelf MLLMs paradigms and can be seamlessly integrated to enhance their efficiency during the VL instruction tuning, which will be quantitatively demonstrated in the following experimental section.

\begin{figure*}[!t]
    \centering
\includegraphics[width=\textwidth,height=\textheight, keepaspectratio]{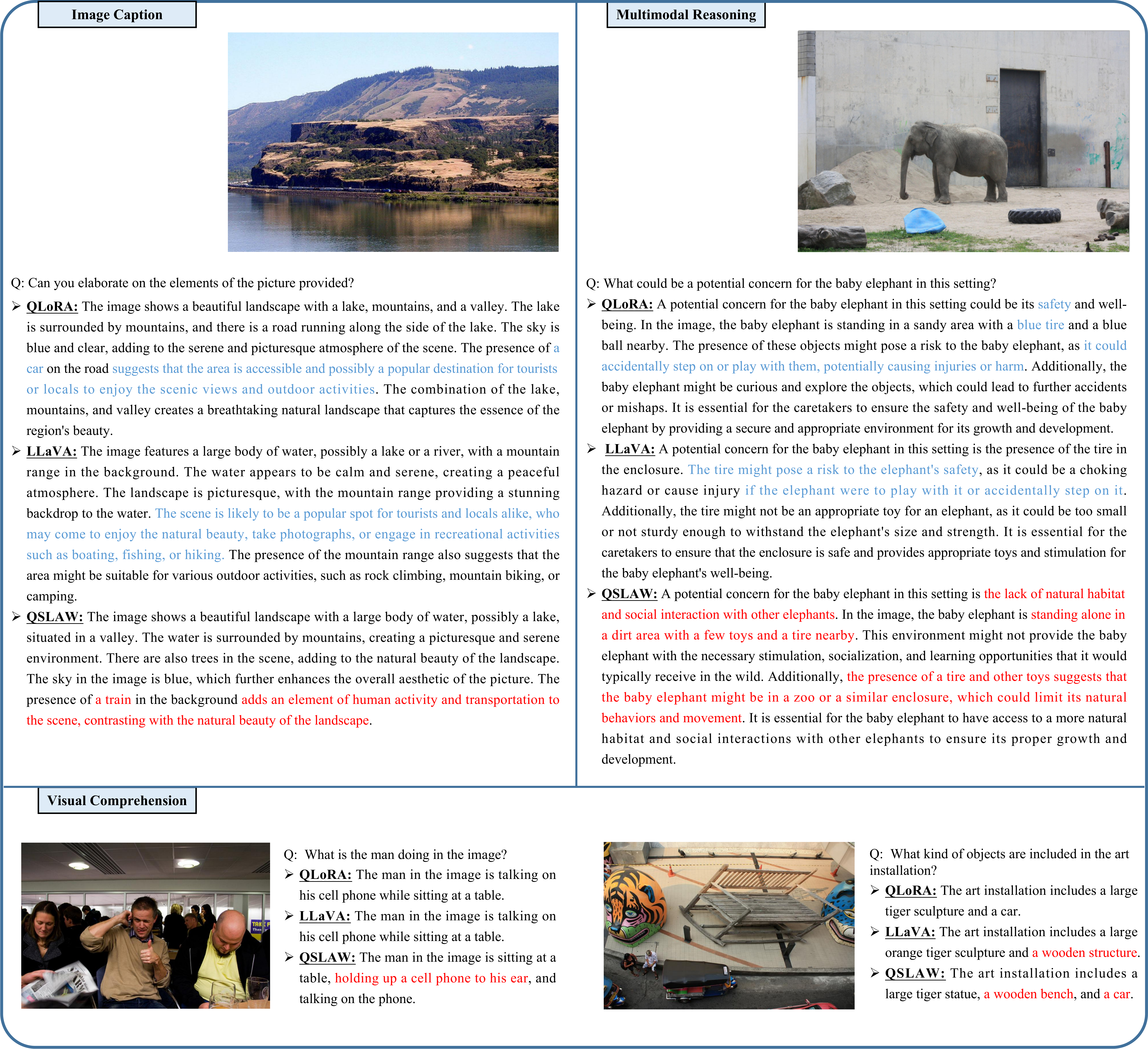}
    \caption{Comparation among various VL instruction tuning paradigms with the examples under different multimodal instruction-following tasks including visual comprehension, image caption and multimodal reasoning. More detailed parts of the response are marked in {\color{red}red} and the misunderstandings in responses are marked in {\color[RGB]{91, 155, 213}bluesky}.}
    \Description{Chat samples.}
    \label{fig:compartion_chat}
\end{figure*}

\section{Experimentation}
\subsection{Experimental Settings}

\subsubsection{Networks and Datasets.} 

To validate the effectiveness of our approach, we select two types of MLLMs: LLaVA~\cite{liu2024visual}, which employs a two-stage full fine-tuning strategy, and LaVIN~\cite{luo2023cheap}, which utilizes one-stage parameter-efficient fine-tuning strategy.
We evaluate performance in line with most multimodal LLMs~\cite{luo2023cheap,liu2024visual,ye2023mplug,dai2024instructblip}, focusing on visual reasoning and instruction-following capabilities.
For a straightforward comparison, we follow the precedent set by LLaVA and LaVIN to choose the ScienceQA dataset~\cite{lu2022learn} for visual reasoning. The dataset, split into \textit{train}, \textit{val}, and \textit{test}, spans diverse domains, including natural science, language science, and social science, and consists of both text-image and text-only inputs.
We report the average accuracy on its \textit{test} split.
For instruction-following, we construct a multimodal ChatBot using LLaVA trained with LLaVA-80k~\cite{liu2024visual}. 
LLaVA-80k is a high-quality vision-language instruction-following dataset generated by ChatGPT/GPT-4~\cite{achiam2023gpt}.
The responses from the ChatBot will be evaluated by GPT-4, with higher-quality responses receiving a score ranging from 1 to 10.

\subsubsection{Implementation Details.}

Following papers~\cite{luo2023cheap,liu2024visual}, we adopt the ViT-L/14 in CLIP as the image encoder.
For LaVIN and LLaVA, we use two MLP layers with a hidden dimension of 128 and a simple linear layer as modular structure, respectively.
For LLMs, we employ LLaMA-7B~\cite{touvron2023llama} and Vicuna-13B~\cite{chiang2023vicuna}.
All parameter settings strictly adhere to the LLaVA and LaVIN papers, except for the 2 training epochs with a batch size of 64, and a 1:1 hybrid training dataset comprising WikiText and downstream data for scale learning.

\begin{figure*}[!t]
    \centering
        \setlength{\abovecaptionskip}{0.4cm}
\setlength{\belowcaptionskip}{-0.cm}
\includegraphics[width=\textwidth,height=\textheight, keepaspectratio]{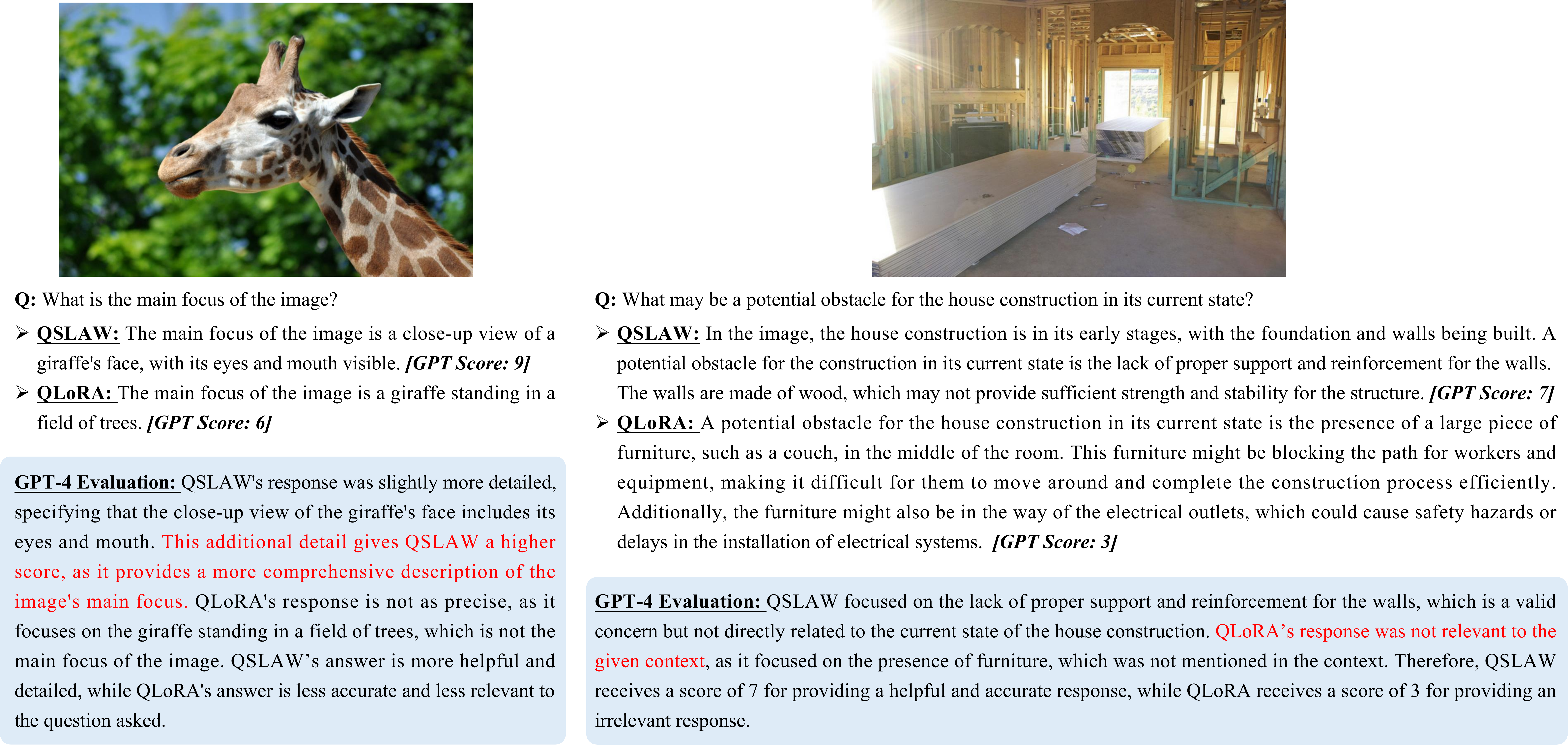}
    \caption{GPT-4 scores for QSLAW and QLoRA. Higher score represents high quality and the reasons why QSLAW obtains a higher score are highlighted in {\color{red}red}.}
    \Description{GPT scores.}
    \label{fig:compartion_gpt}
\end{figure*}

\begin{table*}[t]
    \centering
    \setlength{\abovecaptionskip}{0.3cm}
\setlength{\belowcaptionskip}{-0.cm}
    \caption{Effect of each component on ScienceQA test dataset. All results are conducted with LaVIN-7B.}
    \label{tab:ablation_component}
    \begin{tabular}{lccccccccl}
    \toprule
    \text { Settings }  & \text { NAT } & \text { SOC } & \text { LAN } & \text { TXT } & \text { IMG } & \text { NO } & \text { G1-6 } & \text { G7-12 } & \text { Average (\%) } \\
    \midrule
\text { OmniQuant } & 88.06 & 94.15 & 84.36 & 87.49 & 86.61 & 86.90 & 89.57 & 86.22 & $88.38_{(+0.00)}$ \\
\text { + quantization-aware scaling} & 84.64 & 89.65 & 80.64 & 83.82 & 81.46 & 83.97 & 86.78 & 80.82 & $84.65_{(-3.73)}$ \\
\text { + hybrid data } & 87.26 & 94.83 & 86.64 & 86.56 & 85.77 & 89.41 & 89.98 & 86.35 & $88.68_{(+0.30)}$ \\
\text { + multimodal warmup } & 90.23 & 93.59 & 85.82 & 89.54 & 87.75 & 88.71 & 90.75 & 88.07 & $89.79_{(+1.41)}$ \\
\bottomrule
    \end{tabular}
\end{table*}

\subsection{Main Results}

\subsubsection{ScienceQA}

We categorize MLLMs into one-stage parameter-efficient and two-stage full fine-tuning and select a renowned model for each category to validate our method's performance. Quantitative results on ScienceQA are presented in Table\,\ref{tab:sqa}.
Our approach significantly enhances quantization transfer performance on multimodal tasks, showing consistent improvement across all question classes compared to QLoRA.
For LLaVA, our QSLAW achieves 84.20\% accuracy, a 10.16\% gain over QLoRA's 74.04\%.
With improved settings, our method even outperforms full-precision LLaVA-13B, which fine-tunes the entire LLM on ScienceQA.
For LaVIN, our QSLAW achieves 89.79\% accuracy, a 1.46\% gain compared to QLoRA's 88.33\%, and outperforms full-precision LaVIN-7B. 
%

%

\subsubsection{ChatBot}

We also present qualitative results to demonstrate the multimodal instruction-following capabilities of models obtained using QSLAW.
In Figure\,\ref{fig:compartion_chat}, we compare various VL instruction tuning paradigms with examples from different multimodal instruction-following tasks, including image captioning, multimodal reasoning and visual comprehension.
In more challenging image captioning tasks requiring both local and global image understanding, QSLAW excels.
For a relatively small, distant train in an image, QSLAW correctly identifies it, whereas QLoRA misidentifies it as a car, and LLaVA overlooks it.
This misunderstanding leads LLaVA and QLoRA to generate incorrect speculations about the scene, where it is unlikely for tourists to relax or take photos near a railway.
Additionally, QSLAW exhibits exceptional multimodal reasoning capabilities. It can infer the absence of social interaction from the environment and the position between the tir and the baby elephant, which is contextually consistent.
In contrast, LLaVA and QLoRA merely deduce from the objects in the scene.
For simple visual comprehension questions, QSLAW generates detailed and precise responses.
For instance, QSLAW provides a more comprehensive description of a man's posture while making a phone call compared to QLoRA and LLaVA.
In another image, QSLAW accurately recognizes a wooden structure as a bench and offers a more thorough description of the scene, while QLoRA and LLaVA have omissions.
These examples illustrate that our proposed QSLAW in this paper effectively learns visual knowledge and instruction-following abilities during VL instruction tuning.

We also use very strong GPT-4~\cite{achiam2023gpt} to evaluate the response quality from QSLAW and our QLoRA. The results are reported in Figure\,\ref{fig:compartion_gpt}.
QSLAW performs better than QLoRA, primarily due to its detailed descriptions and superior visual comprehension.

\subsection{Ablation Studies}
\subsubsection{Component Importance}

We examine the effectiveness of each component to provide deeper insights into VL instruction tuning with quantization.
Table\,\ref{tab:ablation_component} shows that when LLM is quantized by OmniQuant and undergoes VL instruction tuning on ScienceQA like LaVIN, it serves as our baseline and achieves higher accuracy compared to LaVIN-QLoRA due to the consideration of activation outliers.
When we introduce quantization-aware scale learning and train it on the same dataset used for VL instruction tuning, the performance drops significantly due to overfitting issues in the LLM backbone.
Incorporating linguistic data to guide scale learning alleviates overfitting and improves average accuracy by 0.66\% compared to the baseline.
Nevertheless, it still lacks effective supervision and exhibits a performance gap compared to full-precision LaVIN (88.54\% for hybrid data \emph{v.s.} 89.41\% for the full-precision).
Our multimodal warmup allows for precise supervision with hybrid data and demonstrates potential beyond models with full-precision VL instruction tuning.

\begin{table}[!t]
\setlength{\abovecaptionskip}{0.3cm}
    \centering
    \caption{Different Quantization initialization method for LoRA and QSLAW. OmniQuant-1 and OmniQuant-2 means the calibration for quantization parameters are conducted on language dataset and hybrid dataset, respectively.}
    \begin{tabular}{cc}
        \toprule
        Initialization & Average (\%)\\
        \midrule
        \rowcolor{black!10}\multicolumn{2}{l}{ \textit{LoRA is used.}}  \\
        NF4 (QLoRA)& 88.27 \\
        OmniQuant-1 & 88.38 \\
        OmniQuant-2 & 88.54 \\
        \midrule
        \rowcolor{black!10}\multicolumn{2}{l}{ \textit{QSLAW is used.}}  \\
        NF4 & 88.98 \\
        OmniQuant-1 & 89.79 \\
        OmniQuant-2 & 89.85 \\
        \bottomrule
    \end{tabular}
    \label{tab:init}
    \vspace{2mm}
\end{table}

\begin{figure}[!t]
    \centering
    \setlength{\abovecaptionskip}{0.2cm}
\setlength{\belowcaptionskip}{-0.cm}
    \includegraphics[width=0.45\textwidth,height=0.3\textheight, keepaspectratio]{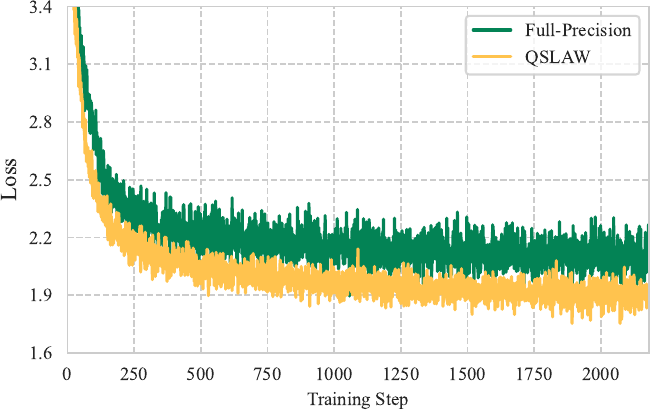}
    \caption{The training process with different strategies. With our multimodal warmup strategy, the training process exhibits faster and more stable fitting.}
    \Description{Pretrain loss}
    \vspace{-1em}
    \label{fig:pretrain_loss}
\end{figure}

\subsubsection{Quantization Initialization}\label{sec:init}

We further evaluate QSLAW's performance with different quantization initialization on ScienceQA.
In Table\,\ref{tab:init}, we examine the performance of LaVIN-7B under both NF4 and OmniQuant.
Our method consistently enhances performance under these two different quantization initializations.
Specifically, QSLAW demonstrates an improvement of 0.71\% and 1.41\% compared to LoRA for NF4 and OmniQuant, respectively.
This result also validates that, for VL tuning where the density and frequency of activation outlier are markedly increased, the NF4 datatype, which aims to equalize the quantity of values across all quantization bins, is sub-optimal and may negatively impact VL tuning.
Moreover, QSLAW outperforms LoRA under both quantization methods, illustrating that our proposed scale learning method is more suitable for VL instruction tuning with quantized LLMs.
This can be attributed to QSLAW's ability to effectively adapt to the unique characteristics of each quantization method, ensuring optimal performance in various quantization scenarios.

In conclusion, QSLAW's versatility and adaptability make it a robust and effective solution for VL instruction tuning across different quantization methods, leading to improved performance and more accurate results in multimodal tasks.


%

%

%

\subsubsection{Alignment Effect}
\begin{figure}[t]
    \setlength{\abovecaptionskip}{0.2cm}
    \setlength{\belowcaptionskip}{-0.2cm}
    \centering
    \includegraphics[width=0.45\textwidth, keepaspectratio]{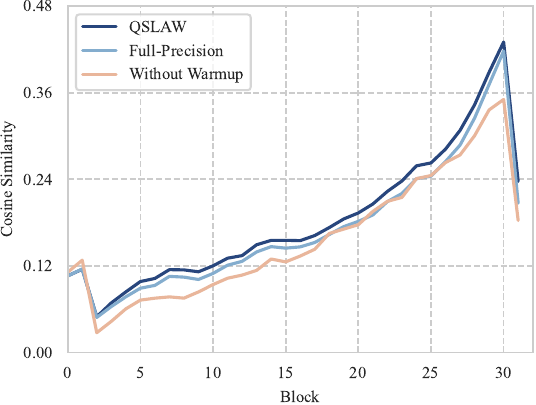}
    \caption{Block-wise cosine similarity between visual tokens and text tokens under different strategies. QSLAW can help model to align visual and textual tokens.}
    \Description{similarity during pre-training for projector}
    \label{fig:similarity}
\end{figure}

To further elucidate the benefits of the proposed QSLAW method in this paper, we conduct in-depth experiments on a two-stage LLaVA model.
This model features a separate stage dedicated to pre-training a modular structure, which allows us to exclude the influence of other trainable parameters.
This setup enables us to observe how quantization-aware scale learning enhances the alignment between visual and language modalities, ultimately leading to improved performance results.
As depicted in Figure\,\ref{fig:pretrain_loss}, QSLAW can stabilize and accelerate the training process for the modular structure. We also calculate the pair-wise cosine similarity between text tokens and image tokens across different layers.
Figure\,\ref{fig:similarity} demonstrates that the modular structure of QSLAW enhances alignment capability, potentially surpassing the projector under full-precision training.
However, such a advantage would be compromised without our multimodal warmup strategy.
This findings highlight the importance of QSLAW's quantization-aware scale learning and multimodal warmup in achieving effective alginment between visual and language modalities.
Such an improved alignment contributes to the model's overall performance and adaptability, making it a valuable approach for multimodal learning tasks.

\section{Conclusion}

In this paper, we are the first to investigate the potential of parameter quantization for MLLMs to reduce training overhead during VL instruction tuning. We propose a Quantization-aware Scale Learning method based on multimodal Warmup (QSLAW).
QSLAW employs quantization and a minimal set of trainable scaling factors to achieve efficient VL instruction tuning. A novel modality-aware warmup is introduced to ensure that scale learning receives adequate multimodal instructional supervision while preserving its original linguistic knowledge.
We validate QSLAW's effectiveness under various settings, demonstrating its excellent multimodal reasoning capabilities. QSLAW surpasses full-precision fine-tuning on ScienceQA and, for ChatBot tasks, effectively learns visual knowledge and instruction-following capabilities.
Our work offers new insights into MLLM quantization and efficient VL instruction tuning, paving the way for further research into exploring the benefits of quantization and constructing more affordable VL instruction tuning methods. We hope this study will inspire additional advancements in the field of multimodal learning and instruction tuning.

\begin{acks}
This work was supported by National Science and Technology Major Project (No. 2022ZD0118201), the National Science Fund for Distinguished Young Scholars (No.62025603), the National Natural Science Foundation of China (No. U21B2037, No. U22B2051, No. U23A20383, No. U21A20472, No. 62176222, No. 62176223, No. 62176226, No. 62072386, No. 62072387, No. 62072389, No. 62002305 and No. 62272401), and the Natural Science Foundation of Fujian Province of China (No.2022J06001).
\end{acks}

%


\bibliographystyle{ACM-Reference-Format}
\bibliography{sample-base}


\begin{thebibliography}{47}


\ifx \showCODEN    \undefined \def \showCODEN     #1{\unskip}     \fi
\ifx \showDOI      \undefined \def \showDOI       #1{#1}\fi
\ifx \showISBNx    \undefined \def \showISBNx     #1{\unskip}     \fi
\ifx \showISBNxiii \undefined \def \showISBNxiii  #1{\unskip}     \fi
\ifx \showISSN     \undefined \def \showISSN      #1{\unskip}     \fi
\ifx \showLCCN     \undefined \def \showLCCN      #1{\unskip}     \fi
\ifx \shownote     \undefined \def \shownote      #1{#1}          \fi
\ifx \showarticletitle \undefined \def \showarticletitle #1{#1}   \fi
\ifx \showURL      \undefined \def \showURL       {\relax}        \fi
\providecommand\bibfield[2]{#2}
\providecommand\bibinfo[2]{#2}
\providecommand\natexlab[1]{#1}
\providecommand\showeprint[2][]{arXiv:#2}

\bibitem[Achiam et~al\mbox{.}(2023)]%
        {achiam2023gpt}
\bibfield{author}{\bibinfo{person}{Josh Achiam}, \bibinfo{person}{Steven Adler}, \bibinfo{person}{Sandhini Agarwal}, \bibinfo{person}{Lama Ahmad}, \bibinfo{person}{Ilge Akkaya}, \bibinfo{person}{Florencia~Leoni Aleman}, \bibinfo{person}{Diogo Almeida}, \bibinfo{person}{Janko Altenschmidt}, \bibinfo{person}{Sam Altman}, \bibinfo{person}{Shyamal Anadkat}, {et~al\mbox{.}}} \bibinfo{year}{2023}\natexlab{}.
\newblock \showarticletitle{Gpt-4 Technical Report}.
\newblock \bibinfo{journal}{\emph{ArXiv}}  \bibinfo{volume}{abs/2303.08774} (\bibinfo{year}{2023}), \bibinfo{numpages}{100}~pages.
\newblock
\urldef\tempurl%
\url{https://api.semanticscholar.org/CorpusID:257532815}
\showURL{%
\tempurl}


\bibitem[Alayrac et~al\mbox{.}(2022)]%
        {alayrac2022flamingo}
\bibfield{author}{\bibinfo{person}{Jean-Baptiste Alayrac}, \bibinfo{person}{Jeff Donahue}, \bibinfo{person}{Pauline Luc}, \bibinfo{person}{Antoine Miech}, \bibinfo{person}{Iain Barr}, \bibinfo{person}{Yana Hasson}, \bibinfo{person}{Karel Lenc}, \bibinfo{person}{Arthur Mensch}, \bibinfo{person}{Katie Millica}, \bibinfo{person}{Roman Ring}, {et~al\mbox{.}}} \bibinfo{year}{2022}\natexlab{}.
\newblock \showarticletitle{Flamingo: A Visual Language Model for Few-Shot Learning}. In \bibinfo{booktitle}{\emph{Advances in neural information processing systems, {NeurIPS} 2022}}, Vol.~\bibinfo{volume}{35}. \bibinfo{publisher}{Curran Associates, Inc.}, \bibinfo{address}{New Orleans}, \bibinfo{pages}{23716--23736}.
\newblock


\bibitem[Anil et~al\mbox{.}(2023)]%
        {team2023gemini}
\bibfield{author}{\bibinfo{person}{Rohan Anil}, \bibinfo{person}{Sebastian Borgeaud}, \bibinfo{person}{Yonghui Wu}, \bibinfo{person}{Jean-Baptiste Alayrac}, \bibinfo{person}{Jiahui Yu}, \bibinfo{person}{Radu Soricut}, \bibinfo{person}{Johan Schalkwyk}, \bibinfo{person}{Andrew~M. Dai}, \bibinfo{person}{Anja Hauth}, {et~al\mbox{.}}} \bibinfo{year}{2023}\natexlab{}.
\newblock \showarticletitle{Gemini: A Family of Highly Capable Multimodal Models}.
\newblock \bibinfo{journal}{\emph{ArXiv}}  \bibinfo{volume}{abs/2312.11805} (\bibinfo{year}{2023}), \bibinfo{numpages}{90}~pages.
\newblock
\urldef\tempurl%
\url{https://api.semanticscholar.org/CorpusID:266361876}
\showURL{%
\tempurl}


\bibitem[Chen et~al\mbox{.}(2020)]%
        {chen2020big}
\bibfield{author}{\bibinfo{person}{Ting Chen}, \bibinfo{person}{Simon Kornblith}, \bibinfo{person}{Kevin Swersky}, \bibinfo{person}{Mohammad Norouzi}, {and} \bibinfo{person}{Geoffrey~E. Hinton}.} \bibinfo{year}{2020}\natexlab{}.
\newblock \showarticletitle{Big Self-supervised Models Are Strong Semi-supervised Learners}. In \bibinfo{booktitle}{\emph{Advances in neural information processing systems, {NeurIPS} 2022}}, Vol.~\bibinfo{volume}{33}. \bibinfo{publisher}{Curran Associates, Inc.}, \bibinfo{address}{New Orleans}, \bibinfo{pages}{22243--22255}.
\newblock


\bibitem[Chen et~al\mbox{.}(2023)]%
        {chen2023pali}
\bibfield{author}{\bibinfo{person}{Xi Chen}, \bibinfo{person}{Josip Djolonga}, \bibinfo{person}{Piotr Padlewski}, \bibinfo{person}{Basil Mustafa}, \bibinfo{person}{Soravit Changpinyo}, \bibinfo{person}{Jialin Wu}, \bibinfo{person}{Carlos~Riquelme Ruiz}, \bibinfo{person}{Sebastian Goodman}, \bibinfo{person}{Xiao Wang}, \bibinfo{person}{Yi Tay}, {et~al\mbox{.}}} \bibinfo{year}{2023}\natexlab{}.
\newblock \bibinfo{title}{Pali-x: On Scaling Up a Multilingual Vision and Language Model}.
\newblock , \bibinfo{numpages}{30}~pages.
\newblock
\urldef\tempurl%
\url{https://api.semanticscholar.org/CorpusID:258967670}
\showURL{%
\tempurl}


\bibitem[Chen et~al\mbox{.}(2024)]%
        {chen2024longlora}
\bibfield{author}{\bibinfo{person}{Yukang Chen}, \bibinfo{person}{Shengju Qian}, \bibinfo{person}{Haotian Tang}, \bibinfo{person}{Xin Lai}, \bibinfo{person}{Zhijian Liu}, \bibinfo{person}{Song Han}, {and} \bibinfo{person}{Jiaya Jia}.} \bibinfo{year}{2024}\natexlab{}.
\newblock \showarticletitle{LongLo{RA}: Efficient Fine-tuning of Long-Context Large Language Models}. In \bibinfo{booktitle}{\emph{International Conference on Learning Representations, {ICLR} 2024}}. \bibinfo{publisher}{OpenReview.net}, \bibinfo{address}{Vienna}, \bibinfo{numpages}{19}~pages.
\newblock
\urldef\tempurl%
\url{https://openreview.net/forum?id=6PmJoRfdaK}
\showURL{%
\tempurl}


\bibitem[Chiang et~al\mbox{.}(2023)]%
        {chiang2023vicuna}
\bibfield{author}{\bibinfo{person}{Wei-Lin Chiang}, \bibinfo{person}{Zhuohan Li}, \bibinfo{person}{Zi Lin}, \bibinfo{person}{Ying Sheng}, \bibinfo{person}{Zhanghao Wu}, \bibinfo{person}{Hao Zhang}, \bibinfo{person}{Lianmin Zheng}, \bibinfo{person}{Siyuan Zhuang}, \bibinfo{person}{Yonghao Zhuang}, \bibinfo{person}{Joseph~E. Gonzalez}, \bibinfo{person}{Ion Stoica}, {and} \bibinfo{person}{Eric~P. Xing}.} \bibinfo{year}{2023}\natexlab{}.
\newblock \bibinfo{title}{Vicuna: An Open-Source Chatbot Impressing GPT-4 with 90\%* ChatGPT Quality}.
\newblock
\newblock
\urldef\tempurl%
\url{https://lmsys.org/blog/2023-03-30-vicuna/}
\showURL{%
\tempurl}


\bibitem[Choi et~al\mbox{.}(2018)]%
        {choi2018pact}
\bibfield{author}{\bibinfo{person}{Jungwook Choi}, \bibinfo{person}{Zhuo Wang}, \bibinfo{person}{Swagath Venkataramani}, \bibinfo{person}{Pierce I-Jen Chuang}, \bibinfo{person}{Vijayalakshmi Srinivasan}, {and} \bibinfo{person}{K. Gopalakrishnan}.} \bibinfo{year}{2018}\natexlab{}.
\newblock \showarticletitle{PACT: Parameterized Clipping Activation for Quantized Neural Networks}.
\newblock \bibinfo{journal}{\emph{ArXiv}}  \bibinfo{volume}{abs/1805.06085} (\bibinfo{year}{2018}), \bibinfo{numpages}{15}~pages.
\newblock
\urldef\tempurl%
\url{https://api.semanticscholar.org/CorpusID:21721698}
\showURL{%
\tempurl}


\bibitem[Chowdhery et~al\mbox{.}(2023)]%
        {chowdhery2023palm}
\bibfield{author}{\bibinfo{person}{Aakanksha Chowdhery}, \bibinfo{person}{Sharan Narang}, \bibinfo{person}{Jacob Devlin}, \bibinfo{person}{Maarten Bosma}, \bibinfo{person}{Gaurav Mishra}, \bibinfo{person}{Adam Roberts}, \bibinfo{person}{Paul Barham}, \bibinfo{person}{Hyung~Won Chung}, \bibinfo{person}{Charles Sutton}, \bibinfo{person}{Sebastian Gehrmann}, {et~al\mbox{.}}} \bibinfo{year}{2023}\natexlab{}.
\newblock \showarticletitle{Palm: Scaling Language Modeling With Pathways}.
\newblock \bibinfo{journal}{\emph{Journal of Machine Learning Research}} \bibinfo{volume}{24}, \bibinfo{number}{240} (\bibinfo{year}{2023}), \bibinfo{pages}{1--113}.
\newblock


\bibitem[Dai et~al\mbox{.}(2023)]%
        {dai2024instructblip}
\bibfield{author}{\bibinfo{person}{Wenliang Dai}, \bibinfo{person}{Junnan Li}, \bibinfo{person}{Dongxu Li}, \bibinfo{person}{Anthony Meng~Huat Tiong}, \bibinfo{person}{Junqi Zhao}, \bibinfo{person}{Weisheng Wang}, \bibinfo{person}{Boyang~Albert Li}, \bibinfo{person}{Pascale Fung}, {and} \bibinfo{person}{Steven C.~H. Hoi}.} \bibinfo{year}{2023}\natexlab{}.
\newblock \showarticletitle{InstructBLIP: Towards General-purpose Vision-Language Models with Instruction Tuning}.
\newblock \bibinfo{journal}{\emph{ArXiv}}  \bibinfo{volume}{abs/2305.06500} (\bibinfo{year}{2023}), \bibinfo{numpages}{17}~pages.
\newblock
\urldef\tempurl%
\url{https://api.semanticscholar.org/CorpusID:258615266}
\showURL{%
\tempurl}


\bibitem[Dettmers et~al\mbox{.}(2022)]%
        {dettmers2022gpt3}
\bibfield{author}{\bibinfo{person}{Tim Dettmers}, \bibinfo{person}{Mike Lewis}, \bibinfo{person}{Younes Belkada}, {and} \bibinfo{person}{Luke Zettlemoyer}.} \bibinfo{year}{2022}\natexlab{}.
\newblock \showarticletitle{Gpt3. int8 (): 8-bit Matrix Multiplication for Transformers at Scale}.
\newblock \bibinfo{journal}{\emph{Advances in Neural Information Processing Systems, {NeurIPS} 2022}}  \bibinfo{volume}{35} (\bibinfo{year}{2022}), \bibinfo{pages}{30318--30332}.
\newblock


\bibitem[Dettmers et~al\mbox{.}(2023)]%
        {dettmers2024qlora}
\bibfield{author}{\bibinfo{person}{Tim Dettmers}, \bibinfo{person}{Artidoro Pagnoni}, \bibinfo{person}{Ari Holtzman}, {and} \bibinfo{person}{Luke Zettlemoyer}.} \bibinfo{year}{2023}\natexlab{}.
\newblock \showarticletitle{QLoRA: Efficient Finetuning of Quantized LLMs}. In \bibinfo{booktitle}{\emph{Advances in Neural Information Processing Systems, {NeurIPS} 2023}}, Vol.~\bibinfo{volume}{36}. \bibinfo{publisher}{Curran Associates, Inc.}, \bibinfo{address}{New Orleans}, \bibinfo{pages}{10088--10115}.
\newblock
\urldef\tempurl%
\url{https://openreview.net/forum?id=OUIFPHEgJU}
\showURL{%
\tempurl}


\bibitem[Dosovitskiy et~al\mbox{.}(2021)]%
        {dosovitskiy2021an}
\bibfield{author}{\bibinfo{person}{Alexey Dosovitskiy}, \bibinfo{person}{Lucas Beyer}, \bibinfo{person}{Alexander Kolesnikov}, \bibinfo{person}{Dirk Weissenborn}, \bibinfo{person}{Xiaohua Zhai}, \bibinfo{person}{Thomas Unterthiner}, \bibinfo{person}{Mostafa Dehghani}, \bibinfo{person}{Matthias Minderer}, \bibinfo{person}{Georg Heigold}, \bibinfo{person}{Sylvain Gelly}, \bibinfo{person}{Jakob Uszkoreit}, {and} \bibinfo{person}{Neil Houlsby}.} \bibinfo{year}{2021}\natexlab{}.
\newblock \showarticletitle{An Image is Worth 16x16 Words: Transformers for Image Recognition at Scale}. In \bibinfo{booktitle}{\emph{International Conference on Learning Representations, {ICLR} 2021}}. \bibinfo{publisher}{OpenReview.net}, \bibinfo{address}{Virtual Only}, \bibinfo{numpages}{21}~pages.
\newblock
\urldef\tempurl%
\url{https://openreview.net/forum?id=YicbFdNTTy}
\showURL{%
\tempurl}


\bibitem[Esser et~al\mbox{.}(2020)]%
        {esser2019learned}
\bibfield{author}{\bibinfo{person}{Steven~K. Esser}, \bibinfo{person}{Jeffrey~L. McKinstry}, \bibinfo{person}{Deepika Bablani}, \bibinfo{person}{Rathinakumar Appuswamy}, {and} \bibinfo{person}{Dharmendra~S. Modha}.} \bibinfo{year}{2020}\natexlab{}.
\newblock \showarticletitle{Learned Step Size Quantization}. In \bibinfo{booktitle}{\emph{International Conference on Learning Representations, {ICLR} 2020}}. \bibinfo{publisher}{OpenReview.net}, \bibinfo{address}{Virtual Only}, \bibinfo{numpages}{12}~pages.
\newblock


\bibitem[Fang et~al\mbox{.}(2023)]%
        {Fang_2023_ICCV}
\bibfield{author}{\bibinfo{person}{Han Fang}, \bibinfo{person}{Zhifei Yang}, \bibinfo{person}{Yuhan Wei}, \bibinfo{person}{Xianghao Zang}, \bibinfo{person}{Chao Ban}, \bibinfo{person}{Zerun Feng}, \bibinfo{person}{Zhongjiang He}, \bibinfo{person}{Yongxiang Li}, {and} \bibinfo{person}{Hao Sun}.} \bibinfo{year}{2023}\natexlab{}.
\newblock \showarticletitle{Alignment and Generation Adapter for Efficient Video-Text Understanding}. In \bibinfo{booktitle}{\emph{Proceedings of the IEEE/CVF International Conference on Computer Vision Workshops, {ICCV workshops} 2023}}. \bibinfo{publisher}{{IEEE}}, \bibinfo{address}{Paris}, \bibinfo{pages}{2791--2797}.
\newblock


\bibitem[Frantar et~al\mbox{.}(2023)]%
        {frantar2023optq}
\bibfield{author}{\bibinfo{person}{Elias Frantar}, \bibinfo{person}{Saleh Ashkboos}, \bibinfo{person}{Torsten Hoefler}, {and} \bibinfo{person}{Dan Alistarh}.} \bibinfo{year}{2023}\natexlab{}.
\newblock \showarticletitle{{OPTQ}: Accurate Quantization for Generative Pre-trained Transformers}. In \bibinfo{booktitle}{\emph{International Conference on Learning Representations, {ICLR} 2023}}. \bibinfo{publisher}{OpenReview.net}, \bibinfo{address}{Kigali Rwanda}, \bibinfo{numpages}{16}~pages.
\newblock
\urldef\tempurl%
\url{https://openreview.net/forum?id=tcbBPnfwxS}
\showURL{%
\tempurl}


\bibitem[Guo et~al\mbox{.}(2024)]%
        {guo2024lqlora}
\bibfield{author}{\bibinfo{person}{Han Guo}, \bibinfo{person}{Philip Greengard}, \bibinfo{person}{Eric Xing}, {and} \bibinfo{person}{Yoon Kim}.} \bibinfo{year}{2024}\natexlab{}.
\newblock \showarticletitle{{LQ}-Lo{RA}: Low-rank plus Quantized Matrix Decomposition for Efficient Language Model Finetuning}. In \bibinfo{booktitle}{\emph{International Conference on Learning Representations, {ICLR} 2024}}. \bibinfo{publisher}{OpenReview.net}, \bibinfo{address}{Vienna}, \bibinfo{numpages}{18}~pages.
\newblock
\urldef\tempurl%
\url{https://openreview.net/forum?id=xw29VvOMmU}
\showURL{%
\tempurl}


\bibitem[Hayou et~al\mbox{.}(2024)]%
        {hayou2024lora+}
\bibfield{author}{\bibinfo{person}{Soufiane Hayou}, \bibinfo{person}{Nikhil Ghosh}, {and} \bibinfo{person}{Bin Yu}.} \bibinfo{year}{2024}\natexlab{}.
\newblock \showarticletitle{LoRA+: Efficient Low Rank Adaptation of Large Models}.
\newblock \bibinfo{journal}{\emph{ArXiv}}  \bibinfo{volume}{abs/2402.12354} (\bibinfo{year}{2024}), \bibinfo{numpages}{27}~pages.
\newblock
\urldef\tempurl%
\url{https://api.semanticscholar.org/CorpusID:267750102}
\showURL{%
\tempurl}


\bibitem[Hendrycks et~al\mbox{.}(2020)]%
        {hendrycks2020measuring}
\bibfield{author}{\bibinfo{person}{Dan Hendrycks}, \bibinfo{person}{Collin Burns}, \bibinfo{person}{Steven Basart}, \bibinfo{person}{Andy Zou}, \bibinfo{person}{Mantas Mazeika}, \bibinfo{person}{Dawn~Xiaodong Song}, {and} \bibinfo{person}{Jacob Steinhardt}.} \bibinfo{year}{2020}\natexlab{}.
\newblock \showarticletitle{Measuring Massive Multitask Language Understanding}. In \bibinfo{booktitle}{\emph{International Conference on Learning Representations, {ICLR} 2021}}. \bibinfo{publisher}{OpenReview.net}, \bibinfo{address}{Virtual Only}, \bibinfo{numpages}{27}~pages.
\newblock


\bibitem[Hu et~al\mbox{.}(2022)]%
        {hu2022lora}
\bibfield{author}{\bibinfo{person}{Edward~J Hu}, \bibinfo{person}{yelong shen}, \bibinfo{person}{Phillip Wallis}, \bibinfo{person}{Zeyuan Allen-Zhu}, \bibinfo{person}{Yuanzhi Li}, \bibinfo{person}{Shean Wang}, \bibinfo{person}{Lu Wang}, {and} \bibinfo{person}{Weizhu Chen}.} \bibinfo{year}{2022}\natexlab{}.
\newblock \showarticletitle{Lo{RA}: Low-Rank Adaptation of Large Language Models}. In \bibinfo{booktitle}{\emph{International Conference on Learning Representations, {ICLR} 2022}}. \bibinfo{publisher}{OpenReview.net}, \bibinfo{address}{Virtual Only}, \bibinfo{numpages}{13}~pages.
\newblock
\urldef\tempurl%
\url{https://openreview.net/forum?id=nZeVKeeFYf9}
\showURL{%
\tempurl}


\bibitem[Jian et~al\mbox{.}(2024)]%
        {jian2024bootstrapping}
\bibfield{author}{\bibinfo{person}{Yiren Jian}, \bibinfo{person}{Chongyang Gao}, {and} \bibinfo{person}{Soroush Vosoughi}.} \bibinfo{year}{2024}\natexlab{}.
\newblock \showarticletitle{Bootstrapping Vision-Language Learning with Decoupled Language Pre-training}. In \bibinfo{booktitle}{\emph{Advances in Neural Information Processing Systems, {NeurIPS} 2023}}, Vol.~\bibinfo{volume}{36}. \bibinfo{publisher}{Curran Associates, Inc.}, \bibinfo{address}{New Orleans}, \bibinfo{pages}{57--72}.
\newblock


\bibitem[Kim et~al\mbox{.}(2024)]%
        {kim2024memory}
\bibfield{author}{\bibinfo{person}{Jeonghoon Kim}, \bibinfo{person}{Jung~Hyun Lee}, \bibinfo{person}{Sungdong Kim}, \bibinfo{person}{Joonsuk Park}, \bibinfo{person}{Kang~Min Yoo}, \bibinfo{person}{Se~Jung Kwon}, {and} \bibinfo{person}{Dongsoo Lee}.} \bibinfo{year}{2024}\natexlab{}.
\newblock \showarticletitle{Memory-efficient Fine-tuning of Compressed Large Language Models via sub-4-bit Integer Quantization}. In \bibinfo{booktitle}{\emph{Advances in Neural Information Processing Systems, {NeurIPS} 2023}}, Vol.~\bibinfo{volume}{36}. \bibinfo{publisher}{Curran Associates, Inc.}, \bibinfo{address}{New Orleans}, \bibinfo{pages}{36187--36207}.
\newblock


\bibitem[Lester et~al\mbox{.}(2021)]%
        {lester2021power}
\bibfield{author}{\bibinfo{person}{Brian Lester}, \bibinfo{person}{Rami Al-Rfou}, {and} \bibinfo{person}{Noah Constant}.} \bibinfo{year}{2021}\natexlab{}.
\newblock \showarticletitle{The Power of Scale for Parameter-Efficient Prompt Tuning}. In \bibinfo{booktitle}{\emph{Empirical Methods in Natural Language Processing, {EMNLP} 2021}}. \bibinfo{publisher}{ACL}, \bibinfo{address}{Punta Cana}, \bibinfo{pages}{3045--3059}.
\newblock
\urldef\tempurl%
\url{https://doi.org/10.18653/V1/2021.EMNLP-MAIN.243}
\showDOI{\tempurl}


\bibitem[Li et~al\mbox{.}(2022)]%
        {li2022blip}
\bibfield{author}{\bibinfo{person}{Junnan Li}, \bibinfo{person}{Dongxu Li}, \bibinfo{person}{Caiming Xiong}, {and} \bibinfo{person}{Steven Hoi}.} \bibinfo{year}{2022}\natexlab{}.
\newblock \showarticletitle{BLIP: Bootstrapping Language-Image Pre-training for Unified Vision-Language Understanding and Generation}. In \bibinfo{booktitle}{\emph{International Conference on Machine Learning, {ICML} 2022}} \emph{(\bibinfo{series}{Proceedings of Machine Learning Research}, Vol.~\bibinfo{volume}{162})}. \bibinfo{publisher}{{PMLR}}, \bibinfo{address}{Baltimore}, \bibinfo{pages}{12888--12900}.
\newblock


\bibitem[Li et~al\mbox{.}(2021)]%
        {li2021brecq}
\bibfield{author}{\bibinfo{person}{Yuhang Li}, \bibinfo{person}{Ruihao Gong}, \bibinfo{person}{Xu Tan}, \bibinfo{person}{Yang Yang}, \bibinfo{person}{Peng Hu}, \bibinfo{person}{Qi Zhang}, \bibinfo{person}{Fengwei Yu}, \bibinfo{person}{Wei Wang}, {and} \bibinfo{person}{Shi Gu}.} \bibinfo{year}{2021}\natexlab{}.
\newblock \showarticletitle{{BRECQ:} Pushing The Limit of Post-Training Quantization by Block Reconstruction}. In \bibinfo{booktitle}{\emph{International Conference on Learning Representations, {ICLR} 2021}}, Vol.~\bibinfo{volume}{162}. \bibinfo{publisher}{OpenReview.net}, \bibinfo{address}{Virtual Only}, \bibinfo{pages}{12888--12900}.
\newblock


\bibitem[Li et~al\mbox{.}(2024)]%
        {li2024loftq}
\bibfield{author}{\bibinfo{person}{Yixiao Li}, \bibinfo{person}{Yifan Yu}, \bibinfo{person}{Chen Liang}, \bibinfo{person}{Nikos Karampatziakis}, \bibinfo{person}{Pengcheng He}, \bibinfo{person}{Weizhu Chen}, {and} \bibinfo{person}{Tuo Zhao}.} \bibinfo{year}{2024}\natexlab{}.
\newblock \showarticletitle{LoftQ: Lo{RA}-Fine-Tuning-aware Quantization for Large Language Models}. In \bibinfo{booktitle}{\emph{International Conference on Learning Representations, {ICLR} 2024}}. \bibinfo{publisher}{OpenReview.net}, \bibinfo{address}{Vienna}, \bibinfo{pages}{16}.
\newblock
\urldef\tempurl%
\url{https://openreview.net/forum?id=LzPWWPAdY4}
\showURL{%
\tempurl}


\bibitem[Lin et~al\mbox{.}(2023)]%
        {lin2023awq}
\bibfield{author}{\bibinfo{person}{Ji Lin}, \bibinfo{person}{Jiaming Tang}, \bibinfo{person}{Haotian Tang}, \bibinfo{person}{Shang Yang}, \bibinfo{person}{Xingyu Dang}, {and} \bibinfo{person}{Song Han}.} \bibinfo{year}{2023}\natexlab{}.
\newblock \showarticletitle{AWQ: Activation-aware Weight Quantization for LLM Compression and Acceleration}.
\newblock \bibinfo{journal}{\emph{ArXiv}}  \bibinfo{volume}{abs/2306.00978} (\bibinfo{year}{2023}), \bibinfo{numpages}{13}~pages.
\newblock
\urldef\tempurl%
\url{https://api.semanticscholar.org/CorpusID:258999941}
\showURL{%
\tempurl}


\bibitem[Liu et~al\mbox{.}(2023a)]%
        {liu2024visual}
\bibfield{author}{\bibinfo{person}{Haotian Liu}, \bibinfo{person}{Chunyuan Li}, \bibinfo{person}{Qingyang Wu}, {and} \bibinfo{person}{Yong~Jae Lee}.} \bibinfo{year}{2023}\natexlab{a}.
\newblock \showarticletitle{Visual Instruction Tuning}. In \bibinfo{booktitle}{\emph{Advances in neural information processing systems, {NeurIPS} 2023}}, Vol.~\bibinfo{volume}{36}. \bibinfo{publisher}{Curran Associates, Inc.}, \bibinfo{address}{New Orleans}, \bibinfo{pages}{34892--34916}.
\newblock


\bibitem[Liu et~al\mbox{.}(2022)]%
        {liu2022nonuniform}
\bibfield{author}{\bibinfo{person}{Zechun Liu}, \bibinfo{person}{Kwang-Ting Cheng}, \bibinfo{person}{Dong Huang}, \bibinfo{person}{Eric~P. Xing}, {and} \bibinfo{person}{Zhiqiang Shen}.} \bibinfo{year}{2022}\natexlab{}.
\newblock \showarticletitle{Nonuniform-to-Uniform Quantization: Towards Accurate Quantization via Generalized Straight-through Estimation}. In \bibinfo{booktitle}{\emph{Proceedings of the IEEE/CVF conference on computer vision and pattern recognition, {CVPR} 2022}}. \bibinfo{publisher}{{IEEE}}, \bibinfo{address}{New Orleans}, \bibinfo{pages}{4942--4952}.
\newblock


\bibitem[Liu et~al\mbox{.}(2023b)]%
        {liu2023llm}
\bibfield{author}{\bibinfo{person}{Zechun Liu}, \bibinfo{person}{Barlas Oğuz}, \bibinfo{person}{Changsheng Zhao}, \bibinfo{person}{Ernie Chang}, \bibinfo{person}{Pierre Stock}, \bibinfo{person}{Yashar Mehdad}, \bibinfo{person}{Yangyang Shi}, \bibinfo{person}{Raghuraman Krishnamoorthi}, {and} \bibinfo{person}{Vikas Chandra}.} \bibinfo{year}{2023}\natexlab{b}.
\newblock \showarticletitle{LLM-QAT: Data-Free Quantization Aware Training for Large Language Models}.
\newblock \bibinfo{journal}{\emph{ArXiv}}  \bibinfo{volume}{abs/2305.17888} (\bibinfo{year}{2023}), \bibinfo{numpages}{15}~pages.
\newblock
\urldef\tempurl%
\url{https://api.semanticscholar.org/CorpusID:258959117}
\showURL{%
\tempurl}


\bibitem[Lu et~al\mbox{.}(2022)]%
        {lu2022learn}
\bibfield{author}{\bibinfo{person}{Pan Lu}, \bibinfo{person}{Swaroop Mishra}, \bibinfo{person}{Tony Xia}, \bibinfo{person}{Liang Qiu}, \bibinfo{person}{Kai-Wei Chang}, \bibinfo{person}{Song-Chun Zhu}, \bibinfo{person}{Oyvind Tafjord}, \bibinfo{person}{Peter Clark}, {and} \bibinfo{person}{A. Kalyan}.} \bibinfo{year}{2022}\natexlab{}.
\newblock \showarticletitle{Learn to Explain: Multimodal Reasoning via Thought Chains for Science Question Answering}.
\newblock \bibinfo{journal}{\emph{Advances in Neural Information Processing Systems, {NeurIPS} 2022}}  \bibinfo{volume}{35} (\bibinfo{year}{2022}), \bibinfo{pages}{2507--2521}.
\newblock


\bibitem[Luo et~al\mbox{.}(2023)]%
        {luo2023cheap}
\bibfield{author}{\bibinfo{person}{Gen Luo}, \bibinfo{person}{Yiyi Zhou}, \bibinfo{person}{Tianhe Ren}, \bibinfo{person}{Shen Chen}, \bibinfo{person}{Xiaoshuai Sun}, {and} \bibinfo{person}{Rongrong Ji}.} \bibinfo{year}{2023}\natexlab{}.
\newblock \showarticletitle{Cheap and Quick: Efficient Vision-Language Instruction Tuning for Large Language Models}. In \bibinfo{booktitle}{\emph{Advances in neural information processing systems，{NeurIPS} 2023}}, Vol.~\bibinfo{volume}{36}. \bibinfo{publisher}{Curran Associates, Inc.}, \bibinfo{address}{New Orleans}, \bibinfo{pages}{29615--29627}.
\newblock


\bibitem[Ma et~al\mbox{.}(2024)]%
        {ma2024affinequant}
\bibfield{author}{\bibinfo{person}{Yuexiao Ma}, \bibinfo{person}{Huixia Li}, \bibinfo{person}{Xiawu Zheng}, \bibinfo{person}{Feng Ling}, \bibinfo{person}{Xuefeng Xiao}, \bibinfo{person}{Rui Wang}, \bibinfo{person}{Shilei Wen}, \bibinfo{person}{Fei Chao}, {and} \bibinfo{person}{Rongrong Ji}.} \bibinfo{year}{2024}\natexlab{}.
\newblock \showarticletitle{AffineQuant: Affine Transformation Quantization for Large Language Models}. In \bibinfo{booktitle}{\emph{International Conference on Learning Representations, {ICLR} 2024}}. \bibinfo{publisher}{OpenReview.net}, \bibinfo{address}{Vienna}, \bibinfo{numpages}{20}~pages.
\newblock
\urldef\tempurl%
\url{https://openreview.net/forum?id=of2rhALq8l}
\showURL{%
\tempurl}


\bibitem[Merity et~al\mbox{.}(2016)]%
        {merity2016pointer}
\bibfield{author}{\bibinfo{person}{Stephen Merity}, \bibinfo{person}{Caiming Xiong}, \bibinfo{person}{James Bradbury}, {and} \bibinfo{person}{Richard Socher}.} \bibinfo{year}{2016}\natexlab{}.
\newblock \showarticletitle{Pointer Sentinel Mixture Models}. In \bibinfo{booktitle}{\emph{International Conference on Learning Representations, {ICLR} 2017}}. \bibinfo{publisher}{OpenReview.net}, \bibinfo{address}{Toulon}, \bibinfo{pages}{13}.
\newblock


\bibitem[Radford et~al\mbox{.}(2019)]%
        {radford2019language}
\bibfield{author}{\bibinfo{person}{Alec Radford}, \bibinfo{person}{Jeff Wu}, \bibinfo{person}{Rewon Child}, \bibinfo{person}{David Luan}, \bibinfo{person}{Dario Amodei}, {and} \bibinfo{person}{Ilya Sutskever}.} \bibinfo{year}{2019}\natexlab{}.
\newblock \bibinfo{title}{Language Models are Unsupervised Multitask Learners}.
\newblock , \bibinfo{numpages}{9}~pages.
\newblock


\bibitem[Raffel et~al\mbox{.}(2020)]%
        {JMLR:v21:20-074}
\bibfield{author}{\bibinfo{person}{Colin Raffel}, \bibinfo{person}{Noam Shazeer}, \bibinfo{person}{Adam Roberts}, \bibinfo{person}{Katherine Lee}, \bibinfo{person}{Sharan Narang}, \bibinfo{person}{Michael Matena}, \bibinfo{person}{Yanqi Zhou}, \bibinfo{person}{Wei Li}, {and} \bibinfo{person}{Peter~J. Liu}.} \bibinfo{year}{2020}\natexlab{}.
\newblock \showarticletitle{Exploring the Limits of Transfer Learning With a Unified Text-to-Text Transformer}.
\newblock \bibinfo{journal}{\emph{Journal of Machine Learning Research}} \bibinfo{volume}{21}, \bibinfo{number}{140} (\bibinfo{year}{2020}), \bibinfo{pages}{1--67}.
\newblock
\urldef\tempurl%
\url{http://jmlr.org/papers/v21/20-074.html}
\showURL{%
\tempurl}


\bibitem[Shao et~al\mbox{.}(2024)]%
        {shao2024omniquant}
\bibfield{author}{\bibinfo{person}{Wenqi Shao}, \bibinfo{person}{Mengzhao Chen}, \bibinfo{person}{Zhaoyang Zhang}, \bibinfo{person}{Peng Xu}, \bibinfo{person}{Lirui Zhao}, \bibinfo{person}{Zhiqian Li}, \bibinfo{person}{Kaipeng Zhang}, \bibinfo{person}{Peng Gao}, \bibinfo{person}{Yu Qiao}, {and} \bibinfo{person}{Ping Luo}.} \bibinfo{year}{2024}\natexlab{}.
\newblock \showarticletitle{OmniQuant: Omnidirectionally Calibrated Quantization for Large Language Models}. In \bibinfo{booktitle}{\emph{International Conference on Learning Representations, {ICLR} 2024}}. \bibinfo{publisher}{OpenReview.net}, \bibinfo{address}{Vienna}, \bibinfo{numpages}{25}~pages.
\newblock
\urldef\tempurl%
\url{https://openreview.net/forum?id=8Wuvhh0LYW}
\showURL{%
\tempurl}


\bibitem[Shin et~al\mbox{.}(2023)]%
        {shin2023nipq}
\bibfield{author}{\bibinfo{person}{Juncheol Shin}, \bibinfo{person}{Junhyuk So}, \bibinfo{person}{Sein Park}, \bibinfo{person}{Seungyeop Kang}, \bibinfo{person}{Sungjoo Yoo}, {and} \bibinfo{person}{Eunhyeok Park}.} \bibinfo{year}{2023}\natexlab{}.
\newblock \showarticletitle{NIPQ: Noise Proxy-based Integrated Pseudo-Quantization}. In \bibinfo{booktitle}{\emph{Proceedings of the IEEE/CVF Conference on Computer Vision and Pattern Recognition, {CVPR} 2023}}. \bibinfo{publisher}{{IEEE}}, \bibinfo{address}{Vancouver}, \bibinfo{pages}{3852--3861}.
\newblock


\bibitem[Touvron et~al\mbox{.}(2023)]%
        {touvron2023llama}
\bibfield{author}{\bibinfo{person}{Hugo Touvron}, \bibinfo{person}{Thibaut Lavril}, \bibinfo{person}{Gautier Izacard}, \bibinfo{person}{Xavier Martinet}, \bibinfo{person}{Marie-Anne Lachaux}, \bibinfo{person}{Timoth{\'e}e Lacroix}, \bibinfo{person}{Baptiste Rozi{\`e}re}, \bibinfo{person}{Naman Goyal}, \bibinfo{person}{Eric Hambro}, \bibinfo{person}{Faisal Azhar}, \bibinfo{person}{Aurelien Rodriguez}, \bibinfo{person}{Armand Joulin}, \bibinfo{person}{Edouard Grave}, {and} \bibinfo{person}{Guillaume Lample}.} \bibinfo{year}{2023}\natexlab{}.
\newblock \showarticletitle{LLaMA: Open and Efficient Foundation Language Models}.
\newblock \bibinfo{journal}{\emph{ArXiv}}  \bibinfo{volume}{abs/2302.13971} (\bibinfo{year}{2023}), \bibinfo{numpages}{27}~pages.
\newblock
\urldef\tempurl%
\url{https://api.semanticscholar.org/CorpusID:257219404}
\showURL{%
\tempurl}


\bibitem[Wang et~al\mbox{.}(2019)]%
        {wang2018glue}
\bibfield{author}{\bibinfo{person}{Alex Wang}, \bibinfo{person}{Amanpreet Singh}, \bibinfo{person}{Julian Michael}, \bibinfo{person}{Felix Hill}, \bibinfo{person}{Omer Levy}, {and} \bibinfo{person}{Samuel~R. Bowman}.} \bibinfo{year}{2019}\natexlab{}.
\newblock \showarticletitle{{GLUE}: A Multi-Task Benchmark and Analysis Platform for Natural Language Understanding}. In \bibinfo{booktitle}{\emph{International Conference on Learning Representations, {ICLR} 2019}}. \bibinfo{publisher}{OpenReview.net}, \bibinfo{address}{New Orleans}, \bibinfo{numpages}{13}~pages.
\newblock
\urldef\tempurl%
\url{https://openreview.net/forum?id=rJ4km2R5t7}
\showURL{%
\tempurl}


\bibitem[Wei et~al\mbox{.}(2022)]%
        {wei2022qdrop}
\bibfield{author}{\bibinfo{person}{Xiuying Wei}, \bibinfo{person}{Ruihao Gong}, \bibinfo{person}{Yuhang Li}, \bibinfo{person}{Xianglong Liu}, {and} \bibinfo{person}{Fengwei Yu}.} \bibinfo{year}{2022}\natexlab{}.
\newblock \showarticletitle{QDrop: Randomly Dropping Quantization for Extremely Low-bit Post-Training Quantization}. In \bibinfo{booktitle}{\emph{International Conference on Learning Representations, {ICLR} 2022}}. \bibinfo{publisher}{OpenReview.net}, \bibinfo{address}{Virtual Only}, \bibinfo{numpages}{19}~pages.
\newblock


\bibitem[Wu et~al\mbox{.}(2023)]%
        {wu2023visual}
\bibfield{author}{\bibinfo{person}{Chenfei Wu}, \bibinfo{person}{Sheng-Kai Yin}, \bibinfo{person}{Weizhen Qi}, \bibinfo{person}{Xiaodong Wang}, \bibinfo{person}{Zecheng Tang}, {and} \bibinfo{person}{Nan Duan}.} \bibinfo{year}{2023}\natexlab{}.
\newblock \showarticletitle{Visual ChatGPT: Talking, Drawing and Editing with Visual Foundation Models}.
\newblock \bibinfo{journal}{\emph{ArXiv}}  \bibinfo{volume}{abs/2303.04671} (\bibinfo{year}{2023}), \bibinfo{numpages}{17}~pages.
\newblock
\urldef\tempurl%
\url{https://api.semanticscholar.org/CorpusID:257404891}
\showURL{%
\tempurl}


\bibitem[Xu et~al\mbox{.}(2024)]%
        {xu2024qalora}
\bibfield{author}{\bibinfo{person}{Yuhui Xu}, \bibinfo{person}{Lingxi Xie}, \bibinfo{person}{Xiaotao Gu}, \bibinfo{person}{Xin Chen}, \bibinfo{person}{Heng Chang}, \bibinfo{person}{Hengheng Zhang}, \bibinfo{person}{Zhengsu Chen}, \bibinfo{person}{XIAOPENG ZHANG}, {and} \bibinfo{person}{Qi Tian}.} \bibinfo{year}{2024}\natexlab{}.
\newblock \showarticletitle{{QA}-Lo{RA}: Quantization-Aware Low-Rank Adaptation of Large Language Models}. In \bibinfo{booktitle}{\emph{International Conference on Learning Representations, {ICLR} 2024}}. \bibinfo{publisher}{OpenReview.net}, \bibinfo{address}{Vienna}, \bibinfo{numpages}{18}~pages.
\newblock
\urldef\tempurl%
\url{https://openreview.net/forum?id=WvFoJccpo8}
\showURL{%
\tempurl}


\bibitem[Ye et~al\mbox{.}(2023)]%
        {ye2023mplug}
\bibfield{author}{\bibinfo{person}{Qinghao Ye}, \bibinfo{person}{Haiyang Xu}, \bibinfo{person}{Guohai Xu}, \bibinfo{person}{Jiabo Ye}, \bibinfo{person}{Ming Yan}, \bibinfo{person}{Yi Zhou}, \bibinfo{person}{Junyan Wang}, \bibinfo{person}{Anwen Hu}, \bibinfo{person}{Pengcheng Shi}, \bibinfo{person}{Yaya Shi}, \bibinfo{person}{Chenliang Li}, \bibinfo{person}{Yuanhong Xu}, \bibinfo{person}{Hehong Chen}, \bibinfo{person}{Junfeng Tian}, \bibinfo{person}{Qiang Qi}, \bibinfo{person}{Ji Zhang}, {and} \bibinfo{person}{Feiyan Huang}.} \bibinfo{year}{2023}\natexlab{}.
\newblock \showarticletitle{mPLUG-Owl: Modularization Empowers Large Language Models with Multimodality}.
\newblock \bibinfo{journal}{\emph{ArXiv}}  \bibinfo{volume}{abs/2304.14178} (\bibinfo{year}{2023}), \bibinfo{numpages}{21}~pages.
\newblock
\urldef\tempurl%
\url{https://api.semanticscholar.org/CorpusID:258352455}
\showURL{%
\tempurl}


\bibitem[Zadouri et~al\mbox{.}(2024)]%
        {zadouri2024pushing}
\bibfield{author}{\bibinfo{person}{Ted Zadouri}, \bibinfo{person}{Ahmet {\"U}st{\"u}n}, \bibinfo{person}{Arash Ahmadian}, \bibinfo{person}{Beyza Ermis}, \bibinfo{person}{Acyr Locatelli}, {and} \bibinfo{person}{Sara Hooker}.} \bibinfo{year}{2024}\natexlab{}.
\newblock \showarticletitle{Pushing Mixture of Experts to the Limit: Extremely Parameter Efficient MoE for Instruction Tuning}. In \bibinfo{booktitle}{\emph{International Conference on Learning Representations, {ICLR} 2024}}. \bibinfo{publisher}{OpenReview.net}, \bibinfo{address}{Vienna}, \bibinfo{numpages}{20}~pages.
\newblock
\urldef\tempurl%
\url{https://openreview.net/forum?id=EvDeiLv7qc}
\showURL{%
\tempurl}


\bibitem[Zhang et~al\mbox{.}(2022)]%
        {zhang2022adaptive}
\bibfield{author}{\bibinfo{person}{Qingru Zhang}, \bibinfo{person}{Minshuo Chen}, \bibinfo{person}{Alexander~W. Bukharin}, \bibinfo{person}{Nikos Karampatziakis}, \bibinfo{person}{Pengcheng He}, \bibinfo{person}{Yu Cheng}, \bibinfo{person}{Weizhu Chen}, {and} \bibinfo{person}{Tuo Zhao}.} \bibinfo{year}{2022}\natexlab{}.
\newblock \showarticletitle{Adaptive Budget Allocation for Parameter-efficient Fine-tuning}. In \bibinfo{booktitle}{\emph{International Conference on Learning Representations, {ICLR} 2023}}. \bibinfo{publisher}{OpenReview.net}, \bibinfo{address}{Kigali}, \bibinfo{numpages}{17}~pages.
\newblock


\bibitem[Zhang et~al\mbox{.}(2024)]%
        {zhang2023llama}
\bibfield{author}{\bibinfo{person}{Renrui Zhang}, \bibinfo{person}{Jiaming Han}, \bibinfo{person}{Chris Liu}, \bibinfo{person}{Aojun Zhou}, \bibinfo{person}{Pan Lu}, \bibinfo{person}{Hongsheng Li}, \bibinfo{person}{Peng Gao}, {and} \bibinfo{person}{Yu Qiao}.} \bibinfo{year}{2024}\natexlab{}.
\newblock \showarticletitle{{LL}a{MA}-Adapter: Efficient Fine-tuning of Large Language Models with Zero-initialized Attention}. In \bibinfo{booktitle}{\emph{International Conference on Learning Representations, {ICLR} 2024}}. \bibinfo{publisher}{OpenReview.net}, \bibinfo{address}{Vienna}, \bibinfo{pages}{22}.
\newblock
\urldef\tempurl%
\url{https://openreview.net/forum?id=d4UiXAHN2W}
\showURL{%
\tempurl}


\end{thebibliography}

\end{document}